# Point-wise or Feature-wise? Benchmark Comparison of Public Available LiDAR Odometry Algorithms in Urban Canyons

Feng Huang, *Student Member, IEEE*, Weisong Wen, Jiachen Zhang and Li-Ta Hsu*, *Member, IEEE*

*Abstract*— **Robust and precise localization is essential for the autonomous system with navigation requirements. Light detection and ranging (LiDAR) odometry is extensively studied in the past decades to achieve this goal. Satisfactory accuracy can be achieved in scenarios with abundant environmental features using existing LiDAR odometry (LO) algorithms. Unfortunately, the performance of the LiDAR odometry is significantly degraded in urban canyons with numerous dynamic objects and complex environmental structures. Meanwhile, it is still not clear from the existing literature which LO algorithms perform well in such challenging environments. To fill this gap, this paper evaluates an array of popular and extensively studied LO pipelines using the datasets collected in urban canyons of Hong Kong. We present the results in terms of their positioning accuracy and computational efficiency. Three major factors dominating the performance of LO in urban canyons are concluded, including the *ego-vehicle dynamic*, *moving objects*, and *degree of urbanization*. According to our experiment results, point-wise achieves better accuracy in urban canyons while feature-wise achieves cost-efficiency and satisfactory positioning accuracy.**

## I. INTRODUCTION

Positioning is one of the major challenges preventing the arrival of fully mobile autonomous systems, such as the autonomous driving vehicle (ADV) [1], autonomous mobile service robots [2], and autonomous aerial robots [3, 4]. The global navigation satellite system (GNSS) is popular for providing globally referenced positioning services. Unfortunately, its performance relies highly on satellite visibility. Satisfactory accuracy can be obtained in open areas with centimeter-level accuracy using GNSS real-time kinematic (RTK). However, the positioning error can be significantly deteriorated in urban canyons with high-rising buildings due to the non-line-of-sight (NLOS) and multipath effects [5]. To mitigate the impacts of the errors caused by multipath and NLOS receptions, numerous researches are proposed to exclude [6, 7], correct [8, 9] the affected GNSS raw measurements to improve the GNSS solution in urban canyons. However, the derived positioning accuracy is still far from the navigation requirements of fully mobile autonomous systems.

The three-dimensions (3D) light detection and ranging (LiDAR), which provides dense 3D point clouds of the surroundings, has recently been widely studied to provide accurate and high-frequency LiDAR odometry (LO) [10] for autonomous systems. The LO method estimates the pose of the autonomous system by accumulating the transformation between consecutive frames of 3D point clouds. Therefore, the accuracy of LO relies heavily on the performance of the point cloud registration, which derives the motion between consecutive frames of 3D point clouds. Numerous methods were studied to perform point cloud registration, which can be mainly divided into feature-wise methods (e.g. Lidar Odometry and Mapping (LOAM) [11]) and pointwise methods (e.g. the Normal Distribution Transform (NDT) [12]). Satisfactory accuracy can be achieved in the widely evaluated KITTI dataset [1] with abundant environmental features and limited moving objects. However, according to our previous finding in [13, 14], NDT-based LO's performance is significantly degraded in challenging urban canyons with numerous moving objects. This is because the existing LO pipelines rely on the assumption that the environmental features are static. In this paper, we aim to present a benchmark of various LO algorithms in the highly urbanized environment for the research community, as well as answering a question: what dominates the performance of LO in dynamic urban canyons. The contributions of this paper are summarized as follows:

(1) This paper presents a comprehensive performance evaluation and analysis of publicly available LO pipelines listed in Table 1 using the challenging datasets collected in typical urban canyons of Hong Kong.
(2) This paper concludes with three dominant factors which degrade the performance of LO algorithms in urban canyons, including the motion difference, dynamic objects, and degree of urbanization.

The rest of the paper is structured as follows: the related works of the existing LO pipelines are reviewed in Section II. Section III presents and compares the theoretical differences between LO pipelines, before the performance evaluation being presented in Section IV. Finally, the conclusions are summarized, and future work is discussed in Section V.

## II. RELATED WORKS

Based on how the raw point clouds are modeled, the available LO methods can be separated into two groups, the point-wise and the feature-wise methods. The point-wise method estimates the relative transformation based on the raw points. Differently, the feature-wise methods [11] extract representative features, such as the edge and planar features from raw point clouds. In other words, the classification of the point-wise and feature-wise methods are based on the assumption that whether all the raw points are directly utilized

Feng Huang, Weisong Wen, Jiachen Zhang, and Li-Ta Hsu, is with Hong Kong Polytechnic University, Hong Kong (e-mail: lt.hsu@polyu.edu.hk).

TABLE 1: Overview of the Selected Publicly-available LO methods evaluated in this paper.

| *Style* | *Method* | *Modeling*[1] | *Optimization*[2] | *Year* |
|---|---|---|---|---|
| Point-wise | *Besl et al, ICP* [15] | minimize the point-to-point distance | s2s, SVD (PCL version) | 1992 |
| | *Segal et al, G-ICP* [16] | surface-based distribution-to-distribution | s2s, BFGS and FLANN (PCL version) | 2009 |
| | *Koide et al, FastGICP* [17] | surface-based distribution-to-distribution | s2s, Gauss-Newton, multi-threaded | 2020 |
| | *Koide et al, VGICP* [17] | voxel-based distribution-to-multi-distribution | s2s, Gauss-Newton, multi-threaded | 2020 |
| | *Koide et al, FastVGICPCuda* [17] | voxel-based distribution-to-multi-distribution | s2s, Gauss-Newton, CUDA-optimized | 2020 |
| | *Biber et al, NDT* [12] | voxel-based point-to-distribution | s2s, Newton method | 2003 |
| | *Koide et al, NDT-OMP* [18] | voxel-based point-to-distribution | s2s, Newton method | 2019 |
| Feature-wise | *Zhang et al, LOAM* [11] | minimize the distance of feature points | s2s and s2m, LM | 2014 |
| | *Qin et al, A-LOAM* [19] | minimize the distance of feature points | s2s and s2m, ceres-solver [20] | 2018 |
| | *Shan et al, LeGO-LOAM* [21] | minimize the distance of feature points, ground-optimization | s2s and s2m, LM | 2018 |
| | *Ye et al, LIO-Mapping* [22] | minimize the distance of feature points | s2s and s2m, LM | 2019 |
| | *Han, Fast LOAM* [23] | minimize the distance of feature points | s2m, ceres-solver [20] | 2020 |

[1]*Modeling* indicates how to model the transformation function.
[2]*Optimization:* s2s, s2m, and PCL represent scan-to-scan, scan-to-map (also called scan-to-model in some references), and Point Cloud Library respectively. SVD, BFGS, FLANN, and LM represent Singular Value Decomposition, Broyden–Fletcher–Goldfarb–Shanno, Fast Library for Approximate Nearest Neighbors, and Levenberg–Marquardt [24] respectively.

in further data association. The publicly available LO methods and their major properties are listed in Table 1. The rest of this section reviews these LO algorithms in detail.

### *A. Point-wise based LiDAR Odometry*

The most well-known point-wise LO method is the iterative closest point (ICP) [15]. ICP is a modular and straightforward algorithm that directly matches two frames of the point cloud by finding the correspondences at a point-wise level. One of the major drawbacks of the ICP is the high computational load arising from the point cloud registration when coping with dense clouds. The performance of the ICP relies on the initial guess due to the non-convexity of the ICP optimization [25]. Moreover, the unexpected outlier measurements arising from moving objects can even introduce additional non-convexity. Many variants of ICP have been proposed to improve its efficiency and accuracy, such as the Trimmed ICP [26] and Normal ICP [27]. Among these ICP variants, the Generalized-ICP (G-ICP) [16] is the most popular variant with distinct accuracy. Instead of formulating the cost function directly by finding point-wise correspondence, the G-ICP, which is benefited from the standard ICP and point-to-plane method introduced by Chen and Medioni [28], optimizes the transformation via a distribution-to-distribution fashion. Compared with the conventional ICP, the G-ICP explores the geometry correlation with points via distribution modeling. As a result, the G-ICP is less sensitive to the initial guess compared with the typical ICP. However, although the G-ICP is known for its accuracy, the optimization of G-ICP shares the same drawback with the ICP and its variants of relying on the time-consuming nearest neighbor search (NNS). According to work in [17], the NNS dominates the computation efficiency of the G-ICP method.

Instead of relying on the time-consuming NNS process, the normal distribution transformation (NDT) [29], which is another major point-wise registration method, employs a voxel-based correspondence association to estimate the transformation. Given a set of raw and discrete point clouds, the NDT divides the point clouds into multiple voxels. Then each voxel is modeled with a Gaussian distribution. The cost function for estimating the transformation is established by mapping the other point clouds to the voxelized one. Overall, the NDT is much faster than the typical ICP. According to an interesting performance comparison work in [30], the NDT outperforms the ICP in terms of the accuracy in the evaluated dataset. However, the performance of the NDT is sensitive to the selection of the resolution of voxels. In other words, carefully tuning the voxel size is needed based on the sensor and environment conditions. Meanwhile, although the NDT does not rely on the time-consuming NNS, the real-time performance still cannot be guaranteed when coping with large amounts of dense point clouds. Fortunately, the team from the Toyohashi University of Technology conducted continuous studies on accelerating the computational efficiency of G-ICP [17] and NDT [18]. The work in [18] proposed and multiple threads accelerated NDT, namely NDT-OMP, for efficient point cloud registration leading to real-time performance. Meanwhile, the continuous work in [17] proposed a voxelized generalized iterative closest point (VGICP) algorithm which is developed on top of the conventional G-ICP, leading to real-time performance. A novel voxelization approach of VGICP enables parallel implementation by aggregating the distribution of all points in a voxel. According to the evaluation of [17], similar or even better performance is obtained using the VGICP compared with the G-ICP.

In short, the G-ICP is one of the most popular and accurate variants. The VGICP relaxed the drawback of the high computation load of the G-ICP arising from the NNS. The NDT is another popular point-wise method and the NDT-OMP enables the real-time implementation of conventional NDT.

### *B. Feature-based LiDAR Odometry*

Instead of directly estimating the transformation using all the raw points, the feature-based LO methods [11, 21, 31, 32] extract a set of representative features from the raw points. The fast point feature histogram (FPFH) was proposed in [31] to extract and describe the features inside the raw points. The FPFH enables the exploration of the local geometry and the transformation is optimized by matching the one-by-one FPFH-based correspondence. A similar descriptor, the histograms of OrienTations (SHOT) is studied in [32] to extract the features. Other works in [33-35] optimized the ground surface features using multiple roadside LiDARs. As the feature-based LO methods only extract a limited set of features, which is significantly fewer than the inputted raw point clouds, the computational load for the matching is significantly lower than the optimization for G-ICP and NDT. Meanwhile, the feature-based method is less sensitive to the initial guess, compared with the point-wise methods. However, it relies on the accuracy of the feature detection and misdetection can lead to incorrect feature association. Theoretically, the feature-based method only makes use of part of the raw points, the convergence accuracy [17] is worse than that of the point-wise LO methods.

Another well-known feature-based LO method is the LOAM [11] which is firstly introduced by the team from Carnegie Mellon University in 2014. Theoretically, the LOAM involves the properties of both the point-wise and the feature-wise method, but we classify the LOAM as the feature-wise group in this paper. On the one hand, to decrease the computation load of the typical ICP, the LOAM proposes to extract two kinds of features, the edge and planar, respectively. The extraction of the feature is simply based on the smoothness of a small region near a given feature point. Different from the FPFH or SHOT which could provide multiple categories of features based on their descriptors, the LOAM only involves two feature groups. Even so, the LOAM shares the advantage of feature-based LO which does not use all the raw points, leading to efficient registration. On the other hand, the LOAM matches the extract edge features to the previously maintained dense edge feature map. Similarly, the extract planar features are matched with the previously maintained dense planar feature map, leading to a scan-to-map matching which is different from the typical ICP (scan-to-scan matching). Interestingly, the matching is evaluated by the Euclidean distance between features, instead of the descriptor's smoothness, which is the same as the point-wise registration. This can significantly decrease the mis-association of features and increase the robustness of the matching. Therefore, the LOAM algorithms share the advantage of low-computation load from the feature-wise method and the robustness of the point-wise method. Due to the superiority of LOAM, it is ranked 2nd according to the evaluation from the KITTI benchmark among the evaluated methods (as of Nov 2020). The average translation error and rotation error is 0.57% and 0.0013 [deg/m] respectively for all possible subsequences of (100, 200, ..., 800) in 3D coordinates [36]. Although the performance evaluation among KITTI dataset is prominent, the accuracy is not guaranteed in the diverse operation scenarios [19, 21, 23]. The work in [21] argues that the LOAM can drift significantly in altitude direction due to the limited features. The LeGO-LOAM is proposed to optimize the altitude direction of the state based on the detected ground points and the drift is slightly decreased in the evaluated dataset. The A-LOAM is proposed in [19] to accelerate the computational efficiency of the LOAM by replacing the complicated derivation of Jacobian with the state-of-the-art non-linear optimization solver, the Ceres-solver [20]. Different from the LOAM, the Fast-LOAM is presented in [23] which is a fast implementation of LOAM. Interestingly, the odometry process is removed in Fast-LOAM and only the mapping process is employed to estimate the transformation. The work in [22] presents an implementation of LOAM, the LIO-Mapping, where the parameters for feature selection and extraction are carefully tuned. However, although the variants of LOAM are studied gradually, the variants did not essentially improve the LOAM. As aforementioned, the LOAM still relies on the availability of the features (edge and planar features). Meanwhile, only part of the raw point clouds is used. As argued by the authors in [17], the combination of the feature-wise LO (e.g LOAM) with the point-wise LO (e.g. VGICP) is a promising solution to guarantee both the accuracy and robustness, where the point-wise fine registration is performed after a coarse feature-wise registration. In short, the LOAM algorithms dominate the existing LO methods since 2015 in the KITTI dataset, due to the following three reasons: (1) edge and planner feature description and their matching guarantee both the real-time performance and accuracy. (2) the scan-to-map matching between features from the keyframe and the historical dense feature map makes it easier to find correspondence accurately.

Recent progress [37, 38] in Neural Network (NN) is another promising solution to improve the performance of the LO methods by deeply exploring the features within point clouds. A study was proposed in [39] to encode the sparse 3D point cloud to dense images then perform the CNN to LO odometry estimation. To tackle the dynamic objects in scenes, LO-Net [40] was proposed with mask-weighted geometric constraint loss which achieved similar results as LOAM. For the generality of the learning-based method, an unsupervised LO method [41] based on the geometric consistency of point clouds is proposed by the team from KAIST in 2020. However, these methods are not open-sourced available and additional training is required thus we do not compare them in this paper.

## C. *Brief Summary*

Both the point-wise and the feature-wise registration methods have their pros and cons. The LO listed in Table 1 will be theoretically compared and experimentally evaluated in the following sections.

TABLE 2: Symbols and their description in this paper

| ***Symbol*** | ***Description*** |
|---|---|
| $N$ | total number of points in a point set |
| Bold Uppercase $\mathbf{X}$ | a set of points, $\mathbf{X}=\{\mathbf{x}_0,\dots,\mathbf{x}_N\}$ |
| $\mathbf{X}_k$ | the point cloud $\mathbf{X}$ received at timestamp $k$ |
| $x, y, z$ | the three-dimensional observation of a point in Cartesian coordinates |
| Bold Lowercase $\mathbf{x}_{(k,i)}$ | the $i^{th}$ point $\mathbf{x}_i$, $\mathbf{x}_i \in \mathbf{X}_k$ |
| $\hat{\mathbf{x}}_{(k,i)}$ | observed value of $\mathbf{x}_{(k,i)}$ |
| $\tilde{\mathbf{x}}_{(k,i)}, \widetilde{\mathbf{X}}_k$ | the $i^{th}$ point or point cloud reconstructed with the initial guess by the previous estimation |
| $\bar{\mathbf{x}}_k, \bar{\mathbf{x}}_k^{voxel}$ | the average of points in $\mathbf{X}_k$ and voxel, respectively |
| $\mathbf{T}_{k+1}^k$ | the estimated transformation $\mathbf{T}$ from the source point cloud $\mathbf{X}_{k+1}$to the target point cloud $\mathbf{X}_k$. $\mathbf{T}_{k+1}^k$ can be divided into translational vector $\mathbf{t}_{k+1}^k$ and rotational motion $\mathbf{R}_{k+1}^k$, $[\mathbf{t}_{k+1}^k, \mathbf{R}_{k+1}^k]= [t_x, t_y, t_z, R_x, R_y, R_z]^T$ |
| $\mathbf{d}_i$ | the residual between corresponding $\mathbf{x}_{(k,i)}$ and $\mathbf{x}_{(k+1,i)}$ |
| $N_i$ | the number of surrounding points around $\mathbf{x}_i$ with a specific distance $r$ |
| $n$ | the number of points in the cell, use in the voxel-based method |
| $\mathbf{C}_{(k,i)}$ | covariance matrix associated with point $\mathbf{x}_{(k,i)}$ |

## III. Theoretical Comparisons of Different LiDAR Odometry Pipelines

In this section, we present the key theory of the listed LO methods. To make a clear comparison, commonly used notations are summarized in Table 2. Matrices are denoted as uppercase bold letters. Vectors are denoted as lowercase bold letters. Variable scalars are denoted as lowercase italic letters. Constant scalars are denoted as lowercase letters. At the end of this section, we summarized their theoretical differences in Table 3.

### A. ICP

The major principle of ICP is to calculate the transformation between the source and target point cloud by minimizing the error of point-to-point distances [15]. Fig. 1 shows an example of registration between two consecutive frames of point clouds. The general cost function of ICP is given below,

$$\mathbf{T}_{k+1}^{k} = \operatorname{argmin}\frac{1}{2}\sum_{i=1}^{N}\left\|\mathbf{x}_{(k,i)} - \left(\mathbf{R}_{k+1}^{k}\mathbf{x}_{(k+1,i)} + \mathbf{t}_{k+1}^{k}\right)\right\|^{2} \tag{1}$$

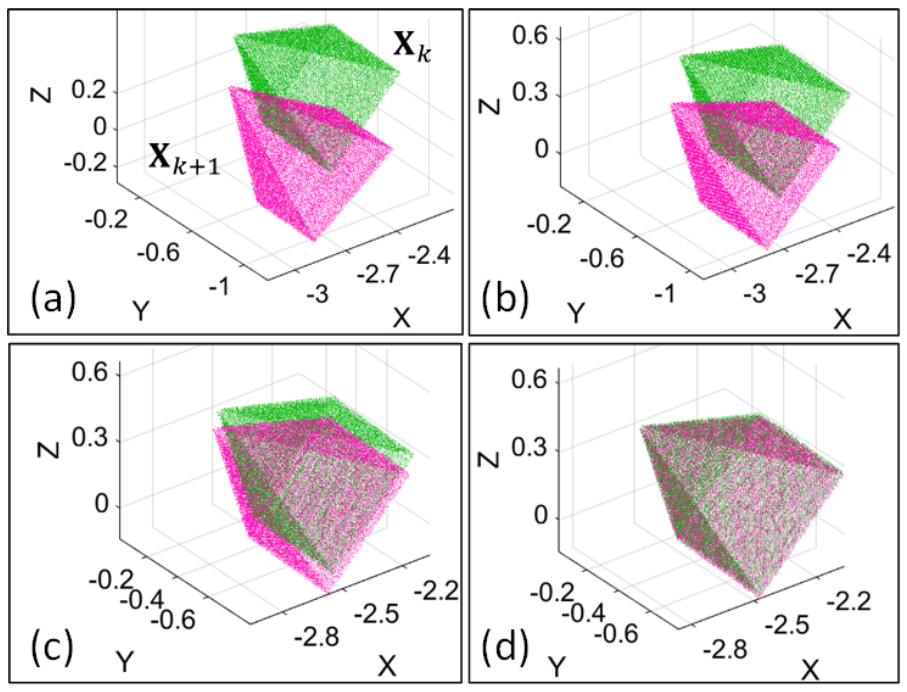

Fig. 1. Point cloud $\mathbf{X}_k$ and $\mathbf{X}_{k+1}$ are from two consecutive scans. (a) represents the original scan. (b) and (c) indicate the transformation after several iterations. (d) shows the results of the final point set registration.

Firstly, the ICP calculates the centroid of the point cloud $\bar{\mathbf{x}}_k$ and $\bar{\mathbf{x}}_{k+1}$.

$$\bar{\mathbf{x}}_{(k)} = \frac{1}{n}\sum_{i=1}^{n}\mathbf{x}_{(k,i)} \tag{2}$$

Then subtract the center of mass from consecutive point sets to obtain the decentering point set $\mathbf{x}'_{(k,i)} \in \mathbf{X}'_k$ and $\mathbf{x}'_{(k+1,i)} \in \mathbf{X}'_{k+1}$ corresponding to $\mathbf{X}_k$ and $\mathbf{X}_{k+1}$, respectively [15],

$$\mathbf{x}'_{(k,i)} = \mathbf{x}_{(k,i)} - \bar{\mathbf{x}}_k, \tag{3}$$

Modifying (1) based on (2) and (3), and we can have,

$$\frac{1}{2}\sum_{i=1}^{N}\left\|\mathbf{x}_{(k,i)} - \left(\mathbf{R}_{k+1}^{k}\mathbf{x}_{(k+1,i)} + \mathbf{t}_{k+1}^{k}\right)\right\|^{2} = \frac{1}{2}\sum_{i=1}^{N}\left\|\mathbf{x}_{(k,i)} - \left(\mathbf{R}_{k+1}^{k}\mathbf{x}_{(k+1,i)} + \mathbf{t}_{k+1}^{k}\right) - \bar{\mathbf{x}}_k + \mathbf{R}_{k+1}^{k}\,\bar{\mathbf{x}}_{k+1} + \bar{\mathbf{x}}_k - \mathbf{R}_{k+1}^{k}\,\bar{\mathbf{x}}_{k+1}\right\|^{2} \tag{4}$$

Then, the right side of (4) can be simplified to,

$$\frac{1}{2}\sum_{i=1}^{N}\left\|\left(\mathbf{x}_{(k,i)} - \bar{\mathbf{x}}_k\right) - \left(\mathbf{R}_{k+1}^{k}\left(\mathbf{x}_{(k+1,i)} - \bar{\mathbf{x}}_{k+1}\right)\right)\right\|^{2} + \left\|\bar{\mathbf{x}}_k - \left(\mathbf{R}_{k+1}^{k}\,\bar{\mathbf{x}}_{k+1} + \mathbf{t}_{k+1}^{k}\right)\right\|^{2} \tag{5}$$

Derive the revised cost function by recalling (4) and (5),

$$\mathbf{T}_{k+1}^{k} = \operatorname{argmin}\frac{1}{2}\sum_{i=1}^{N}\left\|\mathbf{x}'_{(k,i)} - \mathbf{R}\mathbf{x}'_{(k+1,i)}\right\|^{2} + \|(\bar{\mathbf{x}}_k - \mathbf{R}\,\bar{\mathbf{x}}_{k+1} - \mathbf{t}\,\|^{2} \tag{6}$$

Finally, the commonly used singular value decomposition (SVD) [42] is used to compute the estimate translation and rotation iteratively.

### B. Generalized-ICP

Generalized-ICP (G-ICP) [16] extends the classical ICP by combining the ICP and "point-to-plane ICP" into distribution-to-distribution matching. G-ICP assumes each measured point of $\mathbf{X}_k$ and $\mathbf{X}_{k+1}$ can be sampled as a Gaussian distribution $\mathbf{x}_{(k,i)} \sim N(\hat{\mathbf{x}}_{(k,i)}, \mathbf{C}_{(k,i)})$ and $\mathbf{x}_{(k+1,i)} \sim N(\hat{\mathbf{x}}_{(k+1,i)}, \mathbf{C}_{(k+1,i)})$ respectively. $\mathbf{C}_{(k,i)}$ can be calculated from,

$$\mathbf{C}_{(k,i)} = \begin{bmatrix} cov(x,x) & cov(x,y) & cov(x,z) \\ cov(y,x) & cov(y,y) & cov(y,z) \\ cov(z,x) & cov(z,y) & cov(z,z) \end{bmatrix} \tag{7}$$

$$= \begin{bmatrix} E(x^2) - {\mu_x}^2 & E(x,y) - \mu_x\mu_y & E(x,z) - \mu_x\mu_z \\ E(x,y) - \mu_x\mu_y & E(y^2) - {\mu_y}^2 & E(y,z) - \mu_y\mu_z \\ E(x,z) - \mu_x\mu_z & E(y,z) - \mu_y\mu_z & E(z^2) - {\mu_z}^2 \end{bmatrix} \tag{8}$$

$\mu_x, \mu_y$ and $\mu_z$ are the expected value of $\mathbf{x}_{(k,i)}$, approximately equal to the average of its $k$ surrounding points (e.g., k=20 [1] by default in PCL used in k-d tree search). The operator $E(*)$ is used to calculate the expectation of a given component.

Similar to the ICP, the residual of G-ICP can be defined [16] as follows,

$$\hat{\mathbf{d}}_i = \hat{\mathbf{x}}_{(k,i)} - \mathbf{T}_{k+1}\hat{\mathbf{x}}_{(k+1,i)} \tag{9}$$

The residual is assumed to be subjected to Gaussian distribution which describes the geometry correlation between the given point and the neighboring points. Assume $\mathbf{x}_{(k,i)}$ and $\mathbf{x}_{(k+1,i)}$ are independent Gaussian distributed, the distribution of $\mathbf{d}_i$ can be represented as

$$\mathbf{d}_i \sim N\left(\hat{\mathbf{x}}_{(k,i)} - \mathbf{T}_{k+1}\hat{\mathbf{x}}_{(k+1,i)}, \mathbf{C}_{(k,i)} + \mathbf{T}_{k+1}\mathbf{C}_{(k+1,i)}{\mathbf{T}_{k+1}}^{T}\right) \tag{10}$$

$$= N\left(0, \mathbf{C}_{(k,i)} + \mathbf{T}_{k+1}\mathbf{C}_{(k+1,i)}{\mathbf{T}_{k+1}}^{T}\right) \tag{11}$$

Then, we use a maximum likelihood estimation (MLE) to compute the transformation $\mathbf{T}_{k+1}^{k}$ iteratively,

$$\mathbf{T}_{k+1}^{k} = \operatorname{argmax}\sum_{i=1}^{N}\log(p(\mathbf{d}_i)) \tag{12}$$

$$= \operatorname{argmin}\sum_{i=1}^{N}{\mathbf{d}_i}^{T}\left(\mathbf{C}_{(k,i)} + \mathbf{T}_{k+1}^{k}\mathbf{C}_{(k+1,i)}{\mathbf{T}_{k+1}^{k}}^{T}\right)^{-1}\mathbf{d}_i \tag{13}$$

### C. VGICP

Voxelized GICP (VGICP) [17] extends the G-ICP with voxelization to significantly reduce processing time and retain its accuracy as well. First, (15) is extended to compute residual $\mathbf{d}'_i$ between $\mathbf{x}_{(k+1,i)}$ and its neighbor points within distance $r$ in $\mathbf{X}_k$,

$$\{\mathbf{x}_{(k,j)} \mid \|\mathbf{x}_{(k,j)} - \mathbf{x}_{(k+1,i)}\| < r\} \tag{14}$$

$$\hat{\mathbf{d}}'_i = \sum_{j=1}^{N_i} (\hat{\mathbf{x}}_{(k,j)} - \mathbf{T}_{k+1}\hat{\mathbf{x}}_{(k+1,i)}) \tag{15}$$

[1] https://github.com/PointCloudLibrary/pcl/blob/master/registration/include/pcl/registration/gicp.h#L107

Similar to (12) and (13), the distribution of $\mathbf{d}'_i$ can be represented as [17],

$$\mathbf{d}'_i \sim N(0, \sum_{j=1}^{N_i} (\mathbf{C}_{(k,j)} + \mathbf{T}_{k+1}\mathbf{C}_{(k+1,i)}\mathbf{T}_{k+1}{}^T)) \tag{16}$$

where the calculation of the covariance is similar to G-ICP, the cost function can be written as the following.

$$\mathbf{T}^k_{k+1} = \operatorname{argmin} \sum_{i=1}^{N} N_i \left(\frac{\sum \mathbf{x}_{(k,j)}}{N_i} - \mathbf{T}_{k+1}\mathbf{x}_{(k+1,i)}\right)^T \left(\frac{\sum \mathbf{C}_{(k,j)}}{N_i} + \mathbf{T}_{k+1}\mathbf{C}_{(k+1,i)}\mathbf{T}_{k+1}{}^T\right)^{-1} \left(\frac{\sum \mathbf{x}_{(k,j)}}{N_i} - \mathbf{T}_{k+1}\mathbf{x}_{(k+1,i)}\right) \tag{17}$$

The equation above can be further converted into the voxel-based calculation by substituting $\bar{\mathbf{x}}^{voxel}_{(k,i)} = \frac{\sum \mathbf{x}_{(k,j)}}{N_i}$ and $\boldsymbol{C}^{voxel}_{(k,i)} = \frac{\sum \boldsymbol{C}_{(k,j)}}{N_i}$ respectively in each voxel. Finally, the MLE is used to compute the transformation $\mathbf{T}^k_{k+1}$ iteratively as (12) and (13).

Compared with the point correspondence models in ICP or GICP, the data association in VGICP is optimized by exploring the voxel correspondences which is significantly faster than the KD-tree-based neighboring search in ICP. More specifically, if there are ten points in a voxel, the computation load for voxel correspondences could be 10 times smaller than ICP or GICP. As a result, the efficiency of VGICP is significantly faster than the existing point-wise method [16]. However, VGICP needs additional time on KD-tree-based neighboring search for the covariance matrix estimation for each point [17].

## *D. NDT*

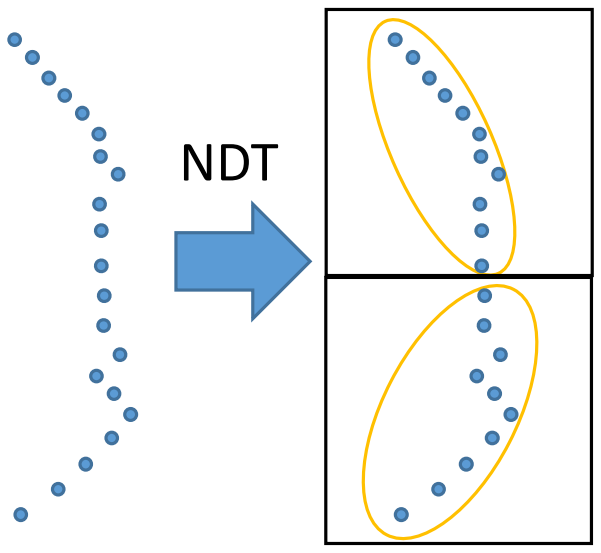

Fig. 2. Points are divided into multiple cells and their distribution in a voxel in 2D-NDT.

Different from the point registration with ICP, the NDT [43] first divided the points cloud into voxels. Then a probability density function (PDF) is computed for each voxel to represent the probability of a point $\mathbf{x}_{(k,i)}$ enclosed by the cell in $\mathbf{X}_k$. The covariance matrix of points inside a cell of $\mathbf{X}_k$,

$$\mathbf{C}^{ndt_cell}_{(k,i)} = \frac{1}{n}\sum_{i=1}^{n} (\mathbf{x}_{(k,i)} - \bar{\mathbf{x}}^{voxel}_{(k,i)})(\mathbf{x}_{(k,i)} - \bar{\mathbf{x}}^{voxel}_{(k,i)})^T) \tag{18}$$

The calculation of the covariance $\boldsymbol{C}^{ndt_cell}_{(k,i)}$ is different from the $\boldsymbol{C}^{voxel}_{(k,i)}$ in VGICP. The $\mathbf{C}^{ndt_cell}_{(k,i)}$ in NDT considers the points in the cell only, at least four or more points in a cell are needed. In other words, a given point is associated with the distribution constructed in the target frame, leading to a point-to-distribution correspondence listed in Table 1. VGICP exploits single-to-multiple distribution by using the average covariance of each point in the same voxel.

The PDF can be formulated as,

$$p(\mathbf{x}_{(k,i)}) = \frac{1}{c}\exp\left(-\frac{(\mathbf{x}_{(k,i)} - \bar{\mathbf{x}}^{voxel}_{(k,i)})^T (\mathbf{C}^{ndt_cell}_{(k,i)})^{-1} (\mathbf{x}_{(k,i)} - \bar{\mathbf{x}}^{voxel}_{(k,i)})}{2}\right) \tag{19}$$

where c is a constant and can be set to one [43]. An example of 2D-NDT is shown in Fig. 2.

The NDT of the scan $\mathbf{X}_k$ is built. Then the maximum matching score function of all points in $\tilde{\mathbf{X}}_{k+1}$ can be converted to minimize the negative sum of the score,

$$\mathbf{T}^k_{k+1} = \operatorname{argmin} \frac{1}{2}\sum_{i=1}^{N} (\tilde{\boldsymbol{x}}_{(k+1,i)} - \bar{\mathbf{x}}^{voxel}_{(k,i)})^T (\mathbf{C}^{ndt_cell}_{(k,i)})^{-1} (\tilde{\boldsymbol{x}}_{(k+1,i)} - \bar{\mathbf{x}}^{voxel}_{(k,i)}) \tag{20}$$

The $\mathbf{T}^k_{k+1}$ can be updated using the Newton method to minimize the score iteratively. Different from the ICP, the NDT employs multiple Gaussian components (each Gaussian component corresponding to a voxel) to describe the geometry of the 3D point clouds. However, the performance of the NDT relies on the selection of the size of the voxel.

NDT-OMP is a modified version of NDT implementation in PCL, which is an OpenMP-boost NDT algorithm with SSE-friendly and multi-threaded [18]. The processing speed of NDT-OMP is up to 10 times faster compared to the original version in PCL.

## *E. LOAM*

LiDAR Odometry and Mapping (LOAM) [11] mainly involves three steps, feature extraction, lidar odometry, and lidar mapping process.

*a) Feature Extraction*

Let $m$ indicates the ring number of point cloud $\mathbf{X}_k$. $\mathbf{S}^m_{(k,i)}$ be a set of continuous neighboring points of $\mathbf{x}_{(k,i)}$ in scan ring $m$. Normally the points in a ring are in clockwise or counterclockwise order according to the receiving time within a scan period (normally 0.1 seconds). The feature points are extracted according to the curvature $c_i$ of point $\mathbf{x}_{(k,i)}$ and its successive points [11],

$$c_i = \frac{1}{N_s * \|\mathbf{x}_{(k,i)}\|} \left\| \sum_{j \in \mathbf{S}^m_{(k,i)}, j \neq i} (\mathbf{x}_{(k,i)} - \mathbf{x}_{(k,j)}) \right\| \tag{21}$$

the $\mathbf{x}_{(k,j)}$ denotes the consecutive point of $\mathbf{x}_{(k,i)}$ within subset $\mathbf{S}^m_{(k,i)}$. $N_s$ represents the number of points in $\mathbf{S}^m_{(k,i)}$, including $\mathbf{x}_{(k,i)}$ and ten consecutive points. The operator $\|\cdot\|$ indicates the L2 vector norm. A point is selected as the edge point if its curvature value is larger than a pre-determined threshold $c_{e-th}$, or classified as a planar point by a smaller curvature in $\mathbf{X}_k$. For points $\mathbf{x}^m_{(k,i)}$ within ring $\mathbf{X}^m_{(k,i)}$ in each scan, we normally divide it into four to eight subregions $\mathbf{X}^{s,m}_{(k,i)}$, and each subregion selects two edge points $\mathbf{x}^e_{(k,i)}$ from $\mathbf{x}^{e,m}_{(k,i)}$ and four planar points $\mathbf{x}^p_{(k,i)}$ within $\mathbf{x}^{p,m}_{(k,i)}$ for odometry process [11], as shown in (22). Ten times edge points and enormous planar features are utilized in mapping process to achieve better accuracy but bring more computational cost. An example of feature extraction results is shown in Fig. 3.

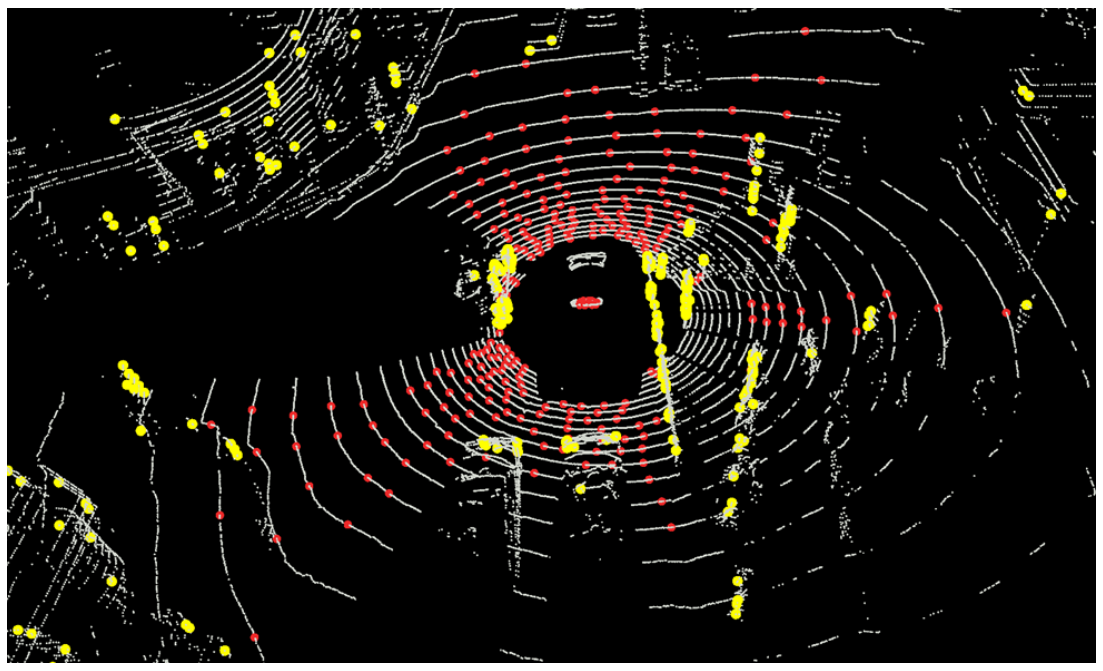

Fig. 3. Illustration of extracted edge (yellow) and planar points (red) for scan-to-map concerning a frame of LiDAR point clouds (grey) (Velodyne HDL-32E) in a crossing road using LOAM.

$$\mathbf{x}^{s,m}_{(k,i)} = \begin{cases} \mathbf{x}^{e}_{(k,i)} \in \mathbf{x}^{e,m}_{(k,i)}, c_i > c_{e-th} \text{ and } N^{e}_{(k,i)} \leq 2 \\ \mathbf{x}^{p}_{(k,i)} \in \mathbf{x}^{p,m}_{(k,i)}, c_i < c_{e-th} \text{ and } N^{p}_{(k,i)} \leq 4 \end{cases} \quad (22)$$

*b) Lidar Odometry*

In general, the odometry is estimated by accumulating the transformation between consecutive frames of point clouds. The role of lidar odometry in LOAM is to estimate the motion between two successive sweeps. The estimated $\mathbf{T}^{L}_{(k,k+1)}$ is used to correct the distortion of points in $\mathbf{X}_{k+1}$ and provide the initial guess to project $\mathbf{X}_k$ as $\widetilde{\mathbf{X}}_k$. During the next frame, the corresponding features can be found between $\widetilde{\mathbf{X}}_k$ and $\mathbf{X}_{k+1}$.

For each edge point $\mathbf{x}^{e}_{(k+1,i)}$, searching its nearest neighbors in $\tilde{\mathbf{x}}_k$ to fit a line by, $\tilde{\mathbf{x}}^{e}_{(k,j)}, \tilde{\mathbf{x}}^{e}_{(k,l)} \in \widetilde{\mathbf{X}}_k$, as the corresponding edge. The distance $d^{e}_{(k+1,i)}$ between the edge point $\mathbf{x}^{e}_{(k+1,i)}$ and the fitted line represents residual of edge feature to be minimized, which can be described as,

$$d^{e}_{(k+1,i)} = \frac{\left|(\mathbf{x}^{e}_{(k+1,i)} - \tilde{\mathbf{x}}^{e}_{(k,j)}) \times (\mathbf{x}^{e}_{(k+1,i)} - \tilde{\mathbf{x}}^{e}_{(k,l)})\right|}{\left|\tilde{\mathbf{x}}^{e}_{(k,j)} - \tilde{\mathbf{x}}^{e}_{(k,l)}\right|} \quad (23)$$

Similarly, for each plane point $\mathbf{x}^{p}_{(k+1,i)}$ in $\mathbf{X}_{k+1}$, the distance $d^{p}_{(k+1,i)}$ between the point and the fitted plane in $\widetilde{\mathbf{X}}_k$, is residual of plane feature to be minimized, which can be represented as,

$$d^{p}_{(k+1,i)} = \frac{\left| \begin{matrix} \left(\mathbf{x}^{p}_{(k+1,i)} - \tilde{\mathbf{x}}^{p}_{(k,j)}\right) \cdot \\ \left|(\tilde{\mathbf{x}}^{p}_{(k,j)} - \tilde{\mathbf{x}}^{p}_{(k,l)}) \times (\tilde{\mathbf{x}}^{p}_{(k,j)} - \tilde{\mathbf{x}}^{p}_{(k,m)})\right| \end{matrix} \right|}{\left|(\tilde{\mathbf{x}}^{p}_{(k,j)} - \tilde{\mathbf{x}}^{p}_{(k,l)}) \times (\tilde{\mathbf{x}}^{p}_{(k,j)} - \tilde{\mathbf{x}}^{p}_{(k,m)})\right|} \quad (24)$$

where $\tilde{\mathbf{x}}^{p}_{(k,j)}, \tilde{\mathbf{x}}^{p}_{(k,l)}, \tilde{\mathbf{x}}^{p}_{(k,m)}$ are three nearest points of $\mathrm{x}^{p}_{(k+1,i)}$ among planar points in $\widetilde{\mathbf{X}}_k$ using k-d tree search.

A geometric relationship between an edge and plane feature point in $\mathbf{X}_{k+1}$ and the corresponding in $\widetilde{\mathbf{X}}_k$ can be estimated as

$$f(\mathbf{x}^{e}_{(k+1,i)}, \mathbf{T}_{k+1}) = d^{e}_{(k+1,i)} \quad (25)$$

$$f(\mathbf{x}^{p}_{(k+1,i)}, \mathbf{T}_{k+1}) = d^{p}_{(k+1,i)} \quad (26)$$

Therefore, the transformation $\mathbf{T}^{k}_{k+1}$ can be calculated by minimizing the cost function based on (25) and (26),

$$\mathbf{T}^{k}_{k+1} = \operatorname{argmin} \frac{1}{2}\left\{\sum_{i=1}^{N^{edge}} \left\|\omega_{(k+1,i)} f\left(\mathbf{x}^{e}_{(k+1,i)}, \mathbf{T}_{k+1}\right)\right\|^2 + \sum_{i=1}^{N^{planar}} \left\|\omega_{(k+1,i)} f\left(\mathbf{x}^{p}_{(k+1,i)}, \mathbf{T}_{k+1}\right)\right\|^2\right\} \quad (27)$$

$N^{edge}$ and $N^{planar}$ are the numbers of edges and planar points obtained in $\mathbf{X}_{k+1}$. Smaller weight $\omega_{(k+1,i)}$ is assigned to feature points with a large residual. The optimization can be solved using the Levenberg-Marquardt [24] method by minimizing the distance of feature.

*c) Lidar Mapping*

The mapping algorithm matches the $\mathbf{X}_{k+1}$ and the point cloud map $\mathbf{M}_k$ to mitigate the error estimation arising from lidar odometry. Let $\mathbf{T}^{L}_{(k,k+1)}$ is the transformation of lidar odometry between $t_k$ and $t_{k+1}$, the initial guess $\mathbf{T}^{W}_{k+1}$ can be represented as

$$\mathbf{T}^{W}_{k+1} = \mathbf{T}^{W}_{k}\mathbf{T}^{L}_{(k,k+1)} \quad (28)$$

Then $\mathbf{T}^{W}_{k+1}$ can be used to transform $\mathbf{X}_{k+1}$ into world coordinate, donated as $\mathbf{X}^{W}_{k+1}$. Similar to the process of lidar odometry in the previous section, the feature-to-feature correspondence can be extracted by applying a k-d tree search between $\mathbf{X}^{W}_{k+1}$ and $\mathbf{M}_k$. Then the $\mathbf{T}^{W}_{k+1}$ can be optimized by minimizing the residuals following (23), (24), and (27). Later the $\mathbf{X}^{W}_{k+1}$ is added to the $\mathbf{M}_k$ to generate the $\mathbf{M}_{k+1}$ for next scan-to-map processing.

Compared with the point-wise theoretically, the curvature is exploited by LOAM for the edges and planar points classification based on (21) and (22). Then the feature points are utilized to compute the inter-frame motion by minimizing the residuals in (23) and (24). Finally, precious state estimation can be obtained with the scan-to-mapping strategy based on the approximate initial guess using (28). Compared with the conventional point-wise LiDAR odometry pipelines, the LOAM extracts the features which can significantly decrease the probability of data mis-association. Moreover, the number of correspondences involved in the optimization is significantly smaller than the one for the point-wise pipeline, leading to improved efficiency.

### *F. LeGO-LOAM*

LeGO-LOAM [21] is a lightweight and ground-optimized LOAM. First, the LeGO-LOAM proposed to project a point cloud onto a range image with a certain resolution base on the horizontal and vertical angular resolution of the LiDAR scanner. Each valid point in the range image has a different pixel value based on the Euclidean distance from the point to the sensor. Then the segmentation is applied to range images to classify the ground and large objects. In the odometry module, the $[\mathrm{t}_z, \theta_{roll}, \theta_{pitch}]$ is obtained by finding the correspondence of planar features from ground points. Meanwhile, the $[\mathrm{t}_x, \mathrm{t}_y, \theta_{yaw}]$ is estimated by matching the edge features from segmented clusters,

$$[t_z, R_x, R_y] = \operatorname{argmin} \frac{1}{2}\sum_{i=1}^{N^{p_ground}} \left\|\omega_{(k+1,i)} f\left(\mathbf{x}^{p}_{(k+1,i)}, \mathbf{T}_{k+1}\right)\right\|^2 \quad (29)$$

$$[t_x, t_y, R_z] = \operatorname{argmin} \frac{1}{2}\sum_{i=1}^{N^{e_seg}} \left\|\omega_{(k+1,i)} f\left(\mathbf{x}^{e}_{(k+1,i)}, \mathbf{T}_{k+1}\right)\right\|^2 \quad (30)$$

where $N^{p_ground}$ and $N^{e_seg}$ are the number of planar pointed within the ground area and edge points from large objects on range image in $\mathbf{X}_{k+1}$. According to the evaluation in [21], the drift in the vertical direction of the LeGO-LOAM significantly smaller than the conventional LOAM since the $[t_z, \theta_{roll}, \theta_{pitch}]$ is innovatively estimated by the ground points. We believe this is one of the major contributions of LeGO-LOAM. Moreover, compared with the conventional LOAM, the LeGO-LOAM also performs a loop closure detection to eliminate drift in the lidar mapping process [44].

## *G. Summary*

The models of the point cloud are different among the diverse methods mentioned above. An overview of the fundamental differences of LO methods selected is shown in Table 3. K-d trees and voxelization are used for point modeling and corresponding search in the LO methods above. G-ICP adopted Gaussian-based point representation hence has the advantage in terms of accuracy and robustness compared to ICP, but with heavily computational cost on the point modeling. NDT is the computational efficient LO method because voxelization is applied for both modeling and corresponding searching. However, the performance is sensitive to the cell size in a diverse environment. VGICP combines the strength of G-ICP and NDT, achieves both fast and accurate point registration.

Feature-wise LO algorithms such as LOAM and its variants employ feature extraction before searching the correspondence, which reduces the computational load thus can be run in real-time for most cases. But the feature-wise method uses a subset of the point cloud and can cause the degeneration case in the scene with limited environmental features. We next move to the evaluation for both point-based and feature-based LO methods towards the challenging of LO methods in the urban canyons.

TABLE 3: Summary of the theoretical differences between various LO methods in terms of the point modeling, data association, and cost function.

| *Method* | *Point Modeling* | *Data Association* | *Cost Function* |
|---|---|---|---|
| *ICP* [15] | Not applicable | k-d tree | $\mathbf{T}_{k+1}^{k} = \operatorname{argmin} \frac{1}{2}\sum_{i=1}^{N} \lVert \mathbf{x}'_{(k,i)} - \mathbf{R}\mathbf{x}'_{(k+1,i)} \rVert^{2} + \lVert (\bar{\mathbf{x}}_k - \mathbf{R}\,\bar{\mathbf{x}}_{k+1} - \mathbf{t} \rVert^{2}$ |
| *G-ICP* [16] | k-d tree | k-d tree | $\mathbf{T}_{k+1}^{k} = \operatorname{argmin} \sum_{i=1}^{N} \mathbf{d}_i{}^{T} (\mathbf{C}_{(k,i)} + \mathbf{T}_{k+1}^{k}\mathbf{C}_{(k+1,i)}\mathbf{T}_{k+1}^{k}{}^{T})^{-1}\mathbf{d}_i$ |
| *VGICP* [17] | k-d tree | voxel | $\mathbf{T}_{k+1}^{k} = \operatorname{argmin} \sum_{i=1}^{N} N_i \left(\frac{\sum \mathbf{x}_{(k,j)}}{N_i} - \mathbf{T}_{k+1}\mathbf{x}_{(k+1,i)}\right)^{T} \left(\frac{\sum \mathbf{C}_{(k,j)}}{N_i} + \mathbf{T}_{k+1}\mathbf{C}_{(k+1,i)}\mathbf{T}_{k+1}{}^{T}\right)^{-1}\left(\frac{\sum \mathbf{x}_{(k,j)}}{N_i} - \mathbf{T}_{k+1}\mathbf{x}_{(k+1,i)}\right)$ |
| *NDT* [12] | voxel | voxel | $\mathbf{T}_{k+1}^{k} = \operatorname{argmin} \frac{1}{2}\sum_{i=1}^{N} \left(\tilde{\boldsymbol{x}}_{(k+1,i)} - \bar{\boldsymbol{x}}_{(k,i)}^{voxel}\right)^{T} (\mathbf{C}_{(k,i)}^{ndt_cell})^{-1}(\tilde{\boldsymbol{x}}_{(k+1,i)} - \bar{\mathbf{x}}_{(k,i)}^{voxel})$ |
| *LOAM* [11] | edge points and planar points | k-d tree | $\mathbf{T}_{k+1}^{k} = \operatorname{argmin} \frac{1}{2}\{\sum_{i=1}^{N^{edge}} \lVert \omega_{(k+1,i)} f\,(\mathbf{x}_{(k+1,i)}^{e}, \mathbf{T}_{k+1}) \rVert^{2} + \sum_{i=1}^{N^{planar}} \lVert \omega_{(k+1,i)} f\,(\boldsymbol{x}_{(k+1,i)}^{e}, \mathbf{T}_{k+1}) \rVert^{2}\}$ |
| *LeGO-LOAM* [21] | clusters edge points and ground planar points | k-d tree | $[t_z, R_x, R_y] = \operatorname{argmin} \frac{1}{2}\sum_{i=1}^{N^{planar}{}_{ground}} \lVert \omega_{(k+1,i)} f\left(\mathbf{x}_{(k+1,i)}^{e}, \mathbf{T}_{k+1}\right) \rVert^{2}$ <br> $[t_x, t_y, R_z] = \operatorname{argmin} \frac{1}{2}\sum_{i=1}^{N^{edge_segmented}} \lVert \omega_{(k+1,i)} f\,(\mathbf{x}_{(k+1,i)}^{p}, \mathbf{T}_{k+1}) \rVert^{2}$ |

TABLE 4: Code implementation and configuration of the publicly available LO methods

| *Method* | *Code Repository* | *Recommended Parameters* | *Rosbag Playback Rate* |
|---|---|---|---|
| *ICP* [15] | github.com/koide3/hdl_graph_slam | Default parameters in hdl_graph_slam.launch | 0.1 |
| *G-ICP* [16] | | | 0.1 |
| *FastGICP* [17] | github.com/SMRT-AIST/fast_gicp | Integrated into hdl_graph_slam. Default parameters are adopted in hdl_graph_slam.launch | 0.5 |
| *FastVGICP* [17] | | | 1.0 |
| *FastVGICPCuda* [17] | | | 1.0 |
| *NDT* [12] | github.com/koide3/hdl_graph_slam | ndt_resolution was set to 3.0 for outdoor environment in hdl_graph_slam.launch | 0.1 |
| *NDT-OMP* [18] | github.com/koide3/ndt_omp/commit/c799f459b4838dd9c65968573370b16c4e7ce7d9 | | 1.0 |
| *LOAM* [11] | github.com/laboshinl/loam_velodyne | Changing VLP-16 to HDL-32 in loam_velodyne.launch. | 1.0 |
| *A-LOAM* [19] | github.com/HKUST-Aerial-Robotics/A-LOAM | Default parameters are adopted in aloam_velodyne_HDL_32.launch. | 1.0 |
| *LeGO-LOAM* [21] | github.com/RobustFieldAutonomyLab/LeGO-LOAM | Changing VLP-16 config to HDL-32E in utility.h | 1.0 |
| *LIO-Mapping* [22] | github.com/hyye/lio-mapping | Updating "sensor_type = 32, no_deskew = false" in 64_scans_test.launch as 32 scans_test.launch | 1.0 |
| *Fast LOAM* [23] | github.com/wh200720041/floam | Changing scan_line = 64 to 32 in floam.launch. | 0.5 |

## IV. Experiments

### A. Experiment Setup

This paper does not use KITTI since it has been widely evaluated in numerous literature and results are summarized in [45]. The performances of the LiDAR odometry methods are evaluated using our recently published UrbanNav dataset [46] which contains the data collected from various degrees of urban areas in Hong Kong and Tokyo. The dataset contains measurements from GNSS, IMU, camera, and LiDAR. Besides, the ground truth data is recorded by NovAtel SPANCPT, which integrates GNSS RTK with fiber optics gyroscope level of IMU. This paper focuses on two competitive datasets, HK-Data20200314 (Data1) and HK-Data20190428 (Data2). Data2 is a more challenging dataset compared with Data1. The typical scenes of both datasets are shown in Fig. 4. The whole datasets are publicly available[2] to the community for further evaluation and algorithm development.

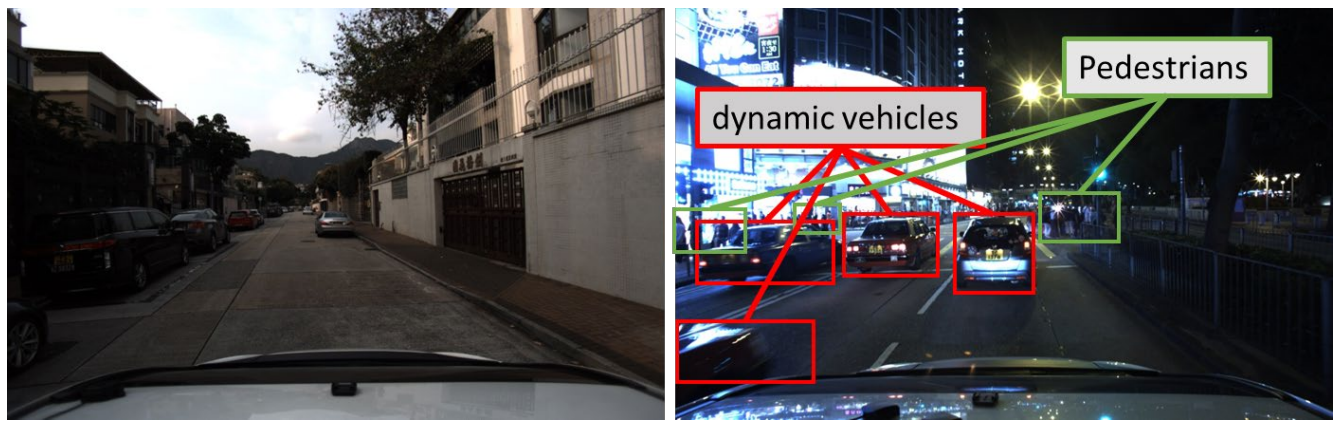

(a) (b)

Fig. 4. Demonstration of the scenarios in the two urban datasets. (a) Data1: Low-rising buildings and multiple right-turning areas in HK-Data20200314. (b) Data2: Variety of dynamic vehicles and numerous high-rising buildings in HK-Data20190428.

The datasets are processed offline with a laptop and the specification of the computing device to evaluate the LiDAR odometry algorithms is provided as below:

- An Intel i7-9750H CPU @ 2.60GHz × 12 threads.
- 2 x 16GB DDR4 2666MHz RAM Memory.
- A GeForce GTX 1660 Ti Graphic Card.

We set up Ubuntu 18.04/ROS melodic as the platform baseline for the majority of the LO pipelines. The original public available LOAM implementation can only be operated in Ubuntu 16.04/ROS Kinetic due to software incompatibilities in the ROS melodic.

The recommended parameters provided by the authors have been retained across all the experiments. We substituted the package configuration for Velodyne HDL32. The detailed configuration and rosbag playback rate are presented in Table 4. For FastGICP and Fast LOAM, the published rate was set to 0.5 as the processing time for some frames is slow than 10Hz. All the experiments are conducted with points data published only. The estimated trajectories of all methods are shown in Fig. 5.

The performance of the LO methods was evaluated via the popular EVO tools[3], a python package that is widely used for evaluating and comparing odometry or SLAM algorithms. We use the relative pose error (RPE) to investigate the local consistency of the SLAM trajectory with standard practice [47]. RPE compares the estimated relative pose with the reference pose in a fixed time interval: Given a sequence of poses from the estimated trajectory $\mathbf{T}_{est} = \{\mathbf{T}_{(est,1)}, \mathbf{T}_{(est,2)}, \dots, \mathbf{T}_{(est,N_{epochs})}\}$ and ground truth $\mathbf{T}_{gt} = \{\mathbf{T}_{(gt,1)}, \mathbf{T}_{(gt,2)}, \dots, \mathbf{T}_{(gt,N_{epochs})}\}$, $N_{epochs}$ represents the number of states for evaluation. The relative pose error $RPE_{i,j}$ between timestamp $t_i$ and $t_j$ can be defined as [48],

$$RPE_{i,j} = (\mathbf{T}_{(gt,i)}^{-1}\mathbf{T}_{(gt,j)})^{-1}(\mathbf{T}_{(est,i)}^{-1}\mathbf{T}_{(est,j)}) \quad (31)$$

$RPE_{i,j}$ belongs to the Special Euclidean Group, $SE(3)$.

$$RMSE = \sqrt{\frac{1}{N_{epochs}}\sum_{i=1}^{N_{epochs}} RPE_i^2} \quad (32)$$

$N_{epochs}$ represents the number of epochs to be evaluated. The RMSE can be calculated based on the overall RPE. The evaluation result of both datasets is shown in Table 6.

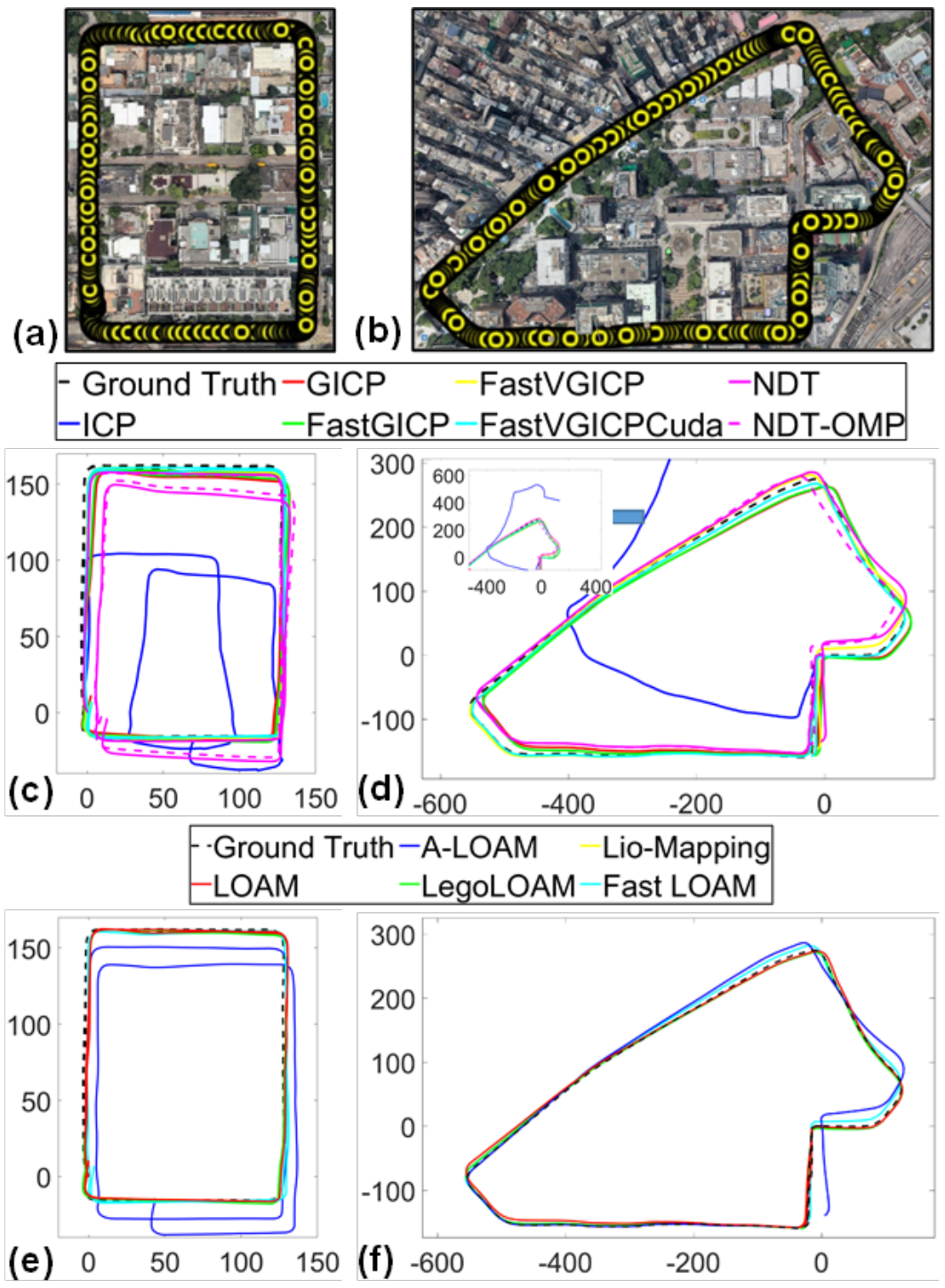

Fig. 5. (a) The satellite image view of ground truth in Data1. (b) The satellite image view of ground truth in Data2. The trajectories in (c)(d) are evaluated by point-wise LO methods of Data1 and Data2, respectively. The trajectories in (e)(f) are evaluated by feature-wise LO methods of Data1 and Data2, respectively.

### B. Computational Cost

The processing time for per-frame (PTPF) is used for the evaluation of the computational cost of LO, as shown in Table 5. The PTPF is measured from receiving a new point cloud to the result of LO odometry. For the point registration-based method, only the processing time of odometry was recorded as PTPF. For LOAM-related methods, the processing time was collected at the end of the odometry and mapping process. LOAM-related methods evaluate the PTPF of odometry and mapping separately, among which LOAM ranked best in the odometry process with a mean value of 9.41 ms while LIO-mapping performed fastest in the mapping process with a mean value of 105.54 ms. Fast LOAM demonstrated a mean

[2] https://www.polyu-ipn-lab.com/download
[3] https://github.com/MichaelGrupp/evo

TABLE 5: The processing time for per-frame (PTPF) evaluation results of Data1 and Data2. Top performance of the registration-based highlighted with **bold** as well as feature-based marked with **blue**. **Red** fonts denote the max of PTPF obviously shifted compare to the mean values.

| *Dataset* | *Method* | *Odometry PTPF (ms)* | | | *Mapping PTPF (ms)* | | |
|---|---|---|---|---|---|---|---|
| | | *Max* | *Min* | *Mean* | *Max* | *Min* | *Mean* |
| Data1 | *ICP* | *1156.76* | *29.99* | *91.47* | | *N/A* | |
| | *G-ICP* | *1476.64* | *92.79* | *232.89* | | *N/A* | |
| | *FastGICP* | *375.21* | *24.50* | *75.02* | | *N/A* | |
| | *FastVGICP* | ***786.19*** | *22.39* | *42.23* | | *N/A* | |
| | *FastVGICPCuda* | ***1182.24*** | ***12.92*** | ***29.00*** | | *N/A* | |
| | *NDT* | *1681.75* | *156.59* | *472.97* | | *N/A* | |
| | *NDT-OMP* | ***301.57*** | *9.35* | *51.67* | | *N/A* | |
| | *LOAM* | *49.04* | ***3.98*** | ***9.41*** | *345.32* | ***16.32*** | *138.81* |
| | *A-LOAM* | *40.74* | *10.73* | *15.93* | *612.79* | *61.27* | *198.51* |
| | *LeGO-LOAM* | ***31.58*** | *4.64* | *9.73* | ***252.27*** | *30.32* | *122.93* |
| | *LIO-Mapping* | *79.88* | *7.88* | *25.22* | *274.58* | *28.94* | ***105.54*** |
| | *Fast LOAM* | | *N/A* | | *124.02* | *22.85* | *60.78* |
| Data2 | *ICP* | *988.06* | *34.42* | *105.38* | | *N/A* | |
| | *G-ICP* | *520.10* | *106.35* | *219.64* | | *N/A* | |
| | *FastGICP* | ***1465.45*** | *21.69* | *87.14* | | *N/A* | |
| | *FastVGICP* | *339.66* | *25.77* | *47.16* | | *N/A* | |
| | *FastVGICPCuda* | ***1626.14*** | ***16.03*** | ***38.08*** | | *N/A* | |
| | *NDT* | *1442.36* | *195.41* | *488.43* | | *N/A* | |
| | *NDT-OMP* | ***142.85*** | *13.87* | *39.18* | | *N/A* | |
| | *LOAM* | *56.85* | *4.7* | ***10.73*** | *322.08* | ***17.50*** | *125.80* |
| | *A-LOAM* | *38.06* | *11.96* | *15.2* | *588.59* | *76.02* | *209.23* |
| | *LeGO-LOAM* | ***33.70*** | ***3.35*** | *11.16* | ***248.47*** | *26.49* | ***115.19*** |
| | *LIO-Mapping* | *86.57* | *9.13* | *28.16* | *328.82* | *37.40* | *119.38* |
| | *Fast LOAM* | | *N/A* | | *164.99* | *27.23* | *90.52* |

PTPF of 60.78ms by merging the odometry and mapping process of the conventional LOAM.

FastVGICPCuda achieved the fastest processing time among the point-registration method, while LOAM performed rapidly in terms of the odometry process. LeGO-LOAM demonstrated the fastest in the mapping process among the LOAM series methods. However, the processing time of the FastGICP, FastVGICP, and FastVGICPCuda had a potential issue when dealing with large data points[4] (marked in red in Table 5) that the max processing speed is much faster than the mean processing speed and published rate in Table 4. The processing delay might result in a large error in some frames.

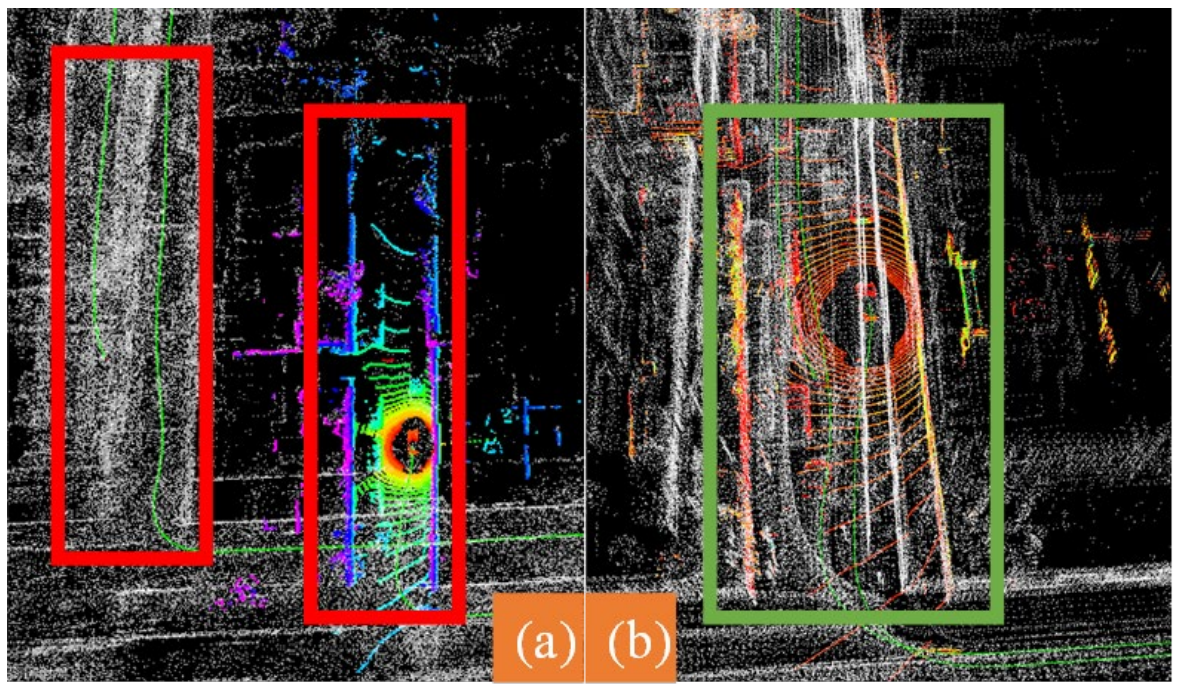

Fig. 6. (a) The mapping process of A-LOAM. The repeating route does not have the overlap which is marked in red (b) replacing the feature extraction process of A-LOAM with the one in LIO-mapping. The pose almost overlays the previous estimation, marked in green color.

## C. Detailed Analysis of Accuracy

Translation error is used as the criteria of accuracy evaluation, as shown in Table 6 and Fig. 7. For Data1, G-ICP obtained the most accurate results among all the evaluated point-registration-related methods, while LOAM achieved the best in all methods in terms of RMSE. For another dataset Data2, G-ICP retained the best performance, while the Fast LOAM performed better than LOAM.

A-LOAM's performance in Table 6 and Fig. 6 (a) of Data1 is much worse than that of the other LOAM-related methods, up to about 0.8 meters in Data1. The reason is that the A-LOAM lacks outlier removal in the feature extraction process, which is part of the original LOAM [11]. To verify this, we perform an experiment to apply the feature extraction part in "LIO-mapping" together with the "A-LOAM" odometry and mapping process. The accuracy of the LiDAR odometry is significantly improved compared with the A-LOAM itself which is shown in Fig. 6 (b) as it can almost overlay when return to start.

In term of the theoretical comparisons in the previous section and the experimental results, we argue that three major factors are affecting the performance of LOs, including (1) the motion difference between two consecutive epochs, which has an impact on the initial guess that project $\mathbf{X}_k$ to $\widetilde{\mathbf{X}}_k$; (2) the density of dynamic objects, a static environment assumption, and the geometry corresponding is affected, and (3) the degree of urbanization of the evaluated environment, which explorer the accuracy of LOs for those highly urbanized areas that GNSS positioning has degenerated [5].

The first factor we considered is the motion difference between the two frames. For calculating the motion difference from the timestamp $t_i$ to $t_{i+1}$, we use the Euclidean norm of the vector $\xi_{k+1}$ in Lie algebra se(3), to represent its corresponding rigid body motion $\mathbf{T}_{k+1}^{k} \in SE(3)$. The formula can be described as,

$$\Delta d_{i+1} = \left\|\log(\mathbf{T}_{k+1}^{k})^{\vee}\right\|, \quad \mathbf{T}_{k+1}^{k} \in R^{4\times 4} \tag{33}$$

where the superscript $\vee$ indicates the de-antisymmetric operation. The operator $\|\cdot\|$ indicates the L2 vector norm.

[4] https://github.com/SMRT-AIST/fast_gicp/issues/17

TABLE 6: Evaluation results on the two urban datasets. The best performance of the registration-based methods in terms of accuracy is highlighted with **bold** as well as that of feature-based ones marked with **blue**.

| *Dataset* | *Description* | *Trajectory Length* | *Method* | *Relative Translation Error (m)* | | *Relative Rotation Error (deg)* | |
|---|---|---|---|---|---|---|---|
| | | | | *RMSE* | *Mean* | *RMSE* | *Mean* |
| Data1 | *Low-urbanization, Small Loop* | *1.21 Km* | *ICP* | *1.857* | *1.529* | *2.073* | *1.533* |
| | | | *G-ICP* | ***0.371*** | *0.326* | *1.912* | *1.268* |
| | | | *FastGICP* | *0.383* | *0.330* | *1.814* | *1.238* |
| | | | *FastVGICP* | *0.372* | ***0.321*** | ***1.670*** | ***1.113*** |
| | | | *FastVGICPCuda* | *0.627* | *0.389* | *1.773* | *1.184* |
| | | | *NDT* | *0.405* | *0.323* | *1.938* | *1.301* |
| | | | *NDT-OMP* | *0.510* | *0.397* | *1.840* | *1.197* |
| | | | *LOAM* | ***0.354*** | ***0.311*** | *2.113* | *1.379* |
| | | | *A-LOAM* | *0.803* | *0.476* | *1.870* | *1.232* |
| | | | *LeGO-LOAM* | *0.374* | *0.324* | *1.972* | *1.236* |
| | | | *LIO-Mapping* | *0.479* | *0.337* | ***1.201*** | ***0.803*** |
| | | | *Fast LOAM* | *0.376* | *0.322* | *1.661* | *1.094* |
| Data2 | *Heavy Traffic, Tall Buildings* | *2.01 Km* | *ICP* | *1.684* | *1.213* | *1.494* | *0.902* |
| | | | *G-ICP* | ***0.417*** | ***0.296*** | *1.129* | *0.662* |
| | | | *FastGICP* | *0.543* | *0.301* | *1.303* | ***0.472*** |
| | | | *FastVGICP* | *0.711* | *0.367* | ***1.093*** | *0.618* |
| | | | *FastVGICPCuda* | *0.874* | *0.410* | *1.734* | *0.732* |
| | | | *NDT* | *0.816* | *0.422* | *1.106* | *0.657* |
| | | | *NDT-OMP* | *0.788* | *0.388* | *1.149* | *0.661* |
| | | | *LOAM* | *0.450* | *0.321* | *1.383* | *0.799* |
| | | | *A-LOAM* | *0.478* | *0.331* | *1.234* | *0.695* |
| | | | *LeGO-LOAM* | *0.462* | *0.333* | *1.226* | *0.694* |
| | | | *LIO-Mapping* | *0.664* | *0.379* | ***0.886*** | ***0.471*** |
| | | | *Fast LOAM* | ***0.423*** | ***0.294*** | *1.141* | *0.619* |

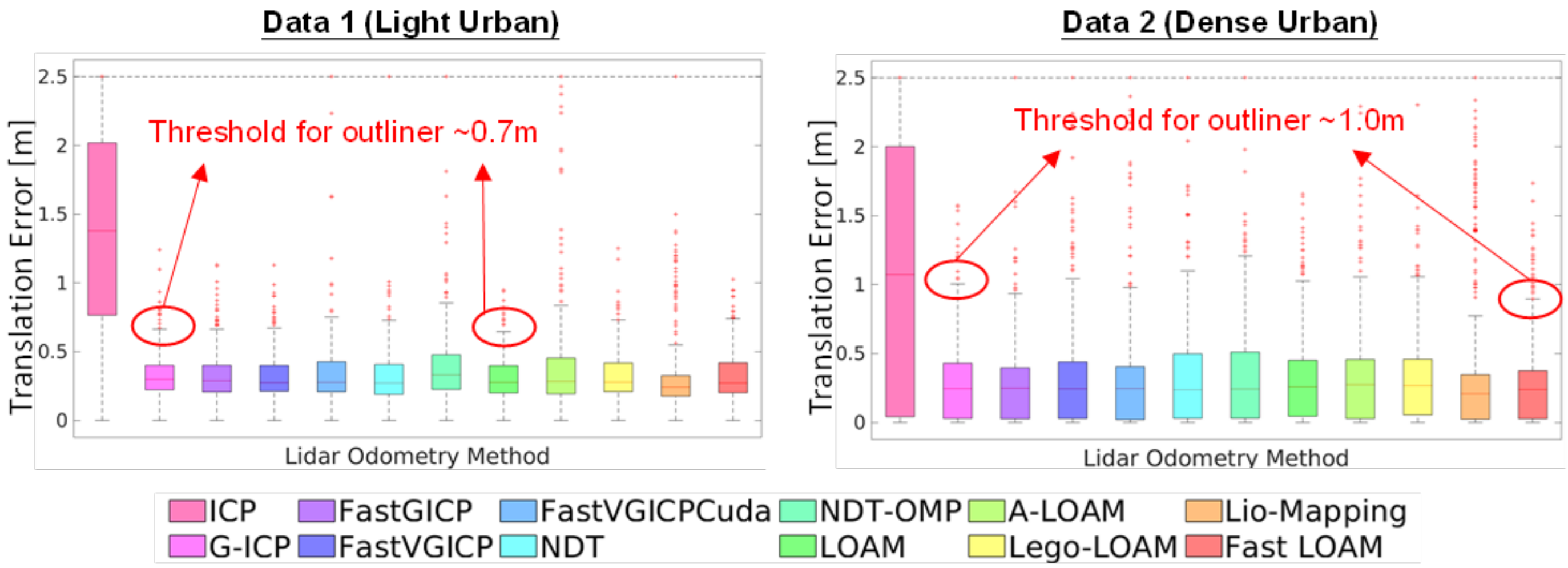

Fig. 7. Boxplot evaluation of the Translation Error Statistics for the LO methods. Whiskers were set as 1.5 times the Interquartile range (IQR) (Q3+1.5*IQR) to better classify the outliers [49]

The second factor is the number of dynamic objects in each scan. We labeled the dynamic objects such as cars and buses in two datasets by SUSTechPOINTS [50], which is a semi-auto annotation tool. The density of dynamic objects factor is defined as,

$$c = \left(\frac{N_{car} + N_{bus}}{N_{total}}\right) \times 100\% \quad (34)$$

which $N_{car}$ and $N_{bus}$ represent the number of LiDAR points of cars and buses separately in the current scan. The $N_{total}$ indicates the total number of points in the current frame.

To evaluate the degree of urbanization of the evaluated scene, our previous work in [51] proposed to adopt the 3D building models to further estimate the skymask (GNSS skyplot with building boundaries). The mean mask elevation angle $\mu_{MEA}$ is defined to quantitatively represent the degree of urbanization as follows:

$$\mu_{MEA} = \frac{\sum_{\alpha=1}^{N} \theta_{\alpha}}{N} \quad (35)$$

where $\theta_{\alpha}$ represented the elevation angle, related to the height of a building, at an azimuth angle $\alpha$, $N$ denotes the number of equally spaced azimuth angles from the skymask, i.e., 360 in the resolution of the azimuth angle is 1 degree. A location with high-rising buildings surrounded result in a large $\mu_{MEA}$ versa vice a small $\mu_{MEA}$ is obtained in rural areas.

To verify the impacts of the listed three factors in the performance of the LO pipelines, we look into the results which are shown in Figs. 8-9. Interquartile range (IQR) [52] is useful to find outliers with a large number of data. The RPE exceeding the Q3+1.5 IQR of the boxplot in Fig. 7 are considered as outliers and marked as a red dot. For example, the threshold to label the outliner of LOs is 0.7 m and 1.0 m for Data1 and Data2, respectively.

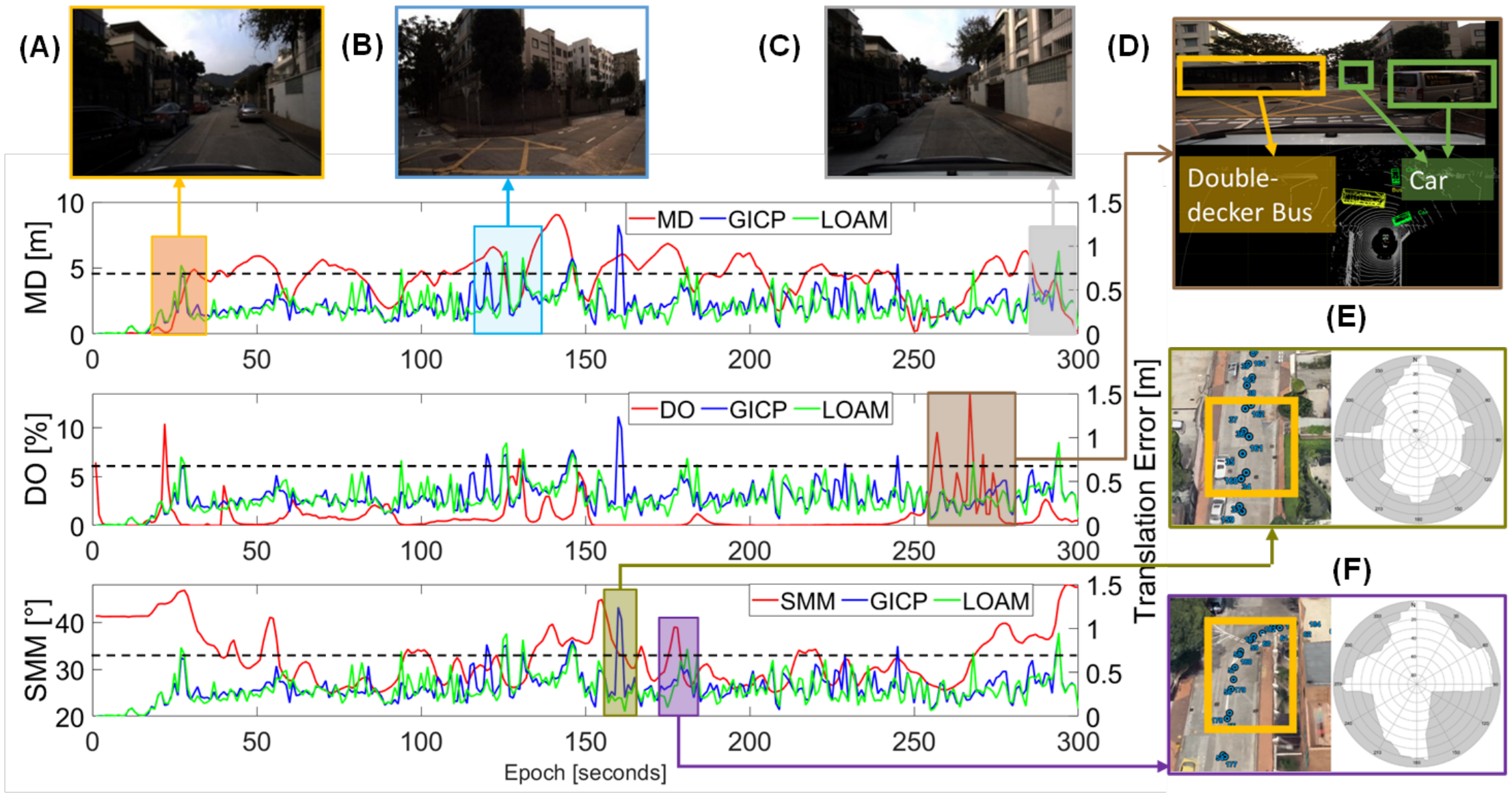

Fig. 8. The comparison of the motion difference (MD), dynamic objects (DO), and sky mask mean (SMM) versus translation error in G-ICP and LOAM using Data1. The dash-dot line indicated the threshold for outliers. (A)(B)(C) are the labels of the typical challenging scenarios for the LO methods.

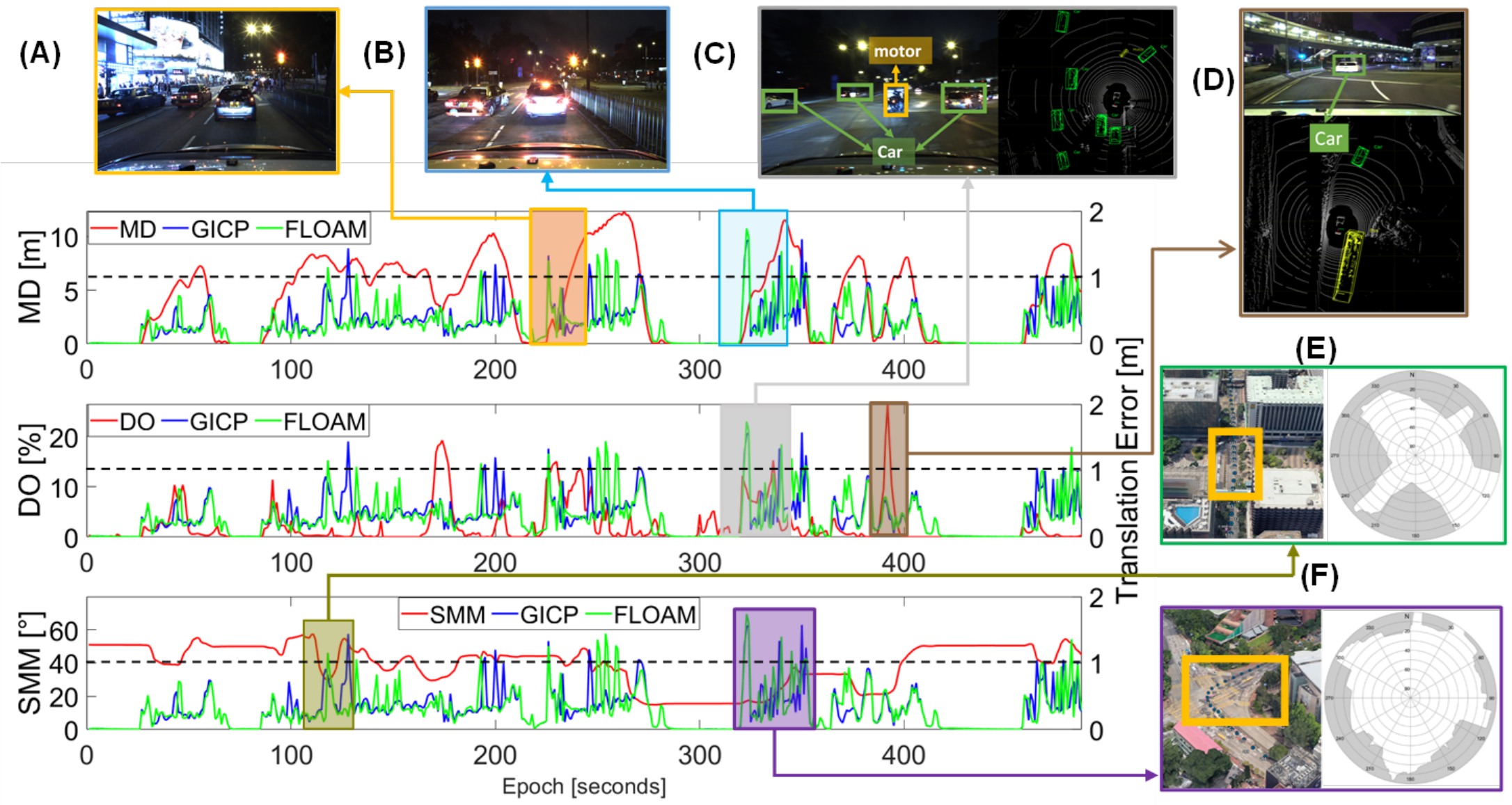

Fig. 9. The comparison of the motion difference (MD), dynamic objects (DO), and sky mask mean (SMM) versus translation error in G-ICP and FLOAM using Data2. The dash-dot line indicated the threshold for outliers. (A)(B)(C) are the labels of the challenging scenarios for the LO methods.

*a) Verification of the listed three factors using Data1*

**Data1 - Factor 1 - Motion Difference:** As shown in Fig. 8 (A), the motion difference increased from 0 to 5.0 m and the translation error jumped to 0.7 m for both G-ICP and LOAM. It is a typical scene of motion changing when a vehicle starts from the roadside. Both methods demonstrated accurate performance between Fig. 8 (A) and (B) as there was a minor change in motion. As shown in Fig. 8 (B), the ego-vehicle slows down before passing the crossing and thus produces a significant motion offset. As shown in Fig. 8 (C), the error increases again as the car turned right, then returned to the start point and stopped at the end finally.

**Data1 -Factor 2 - Dynamic Objects:** Data1 is a low traffic area without any vehicles excepting (D) in Fig. 8. The results show that the LOAM is more susceptible to dynamic vehicles compared with the G-ICP in the evaluated dataset.

**Data1 -Factor 3 - Degree of the Urbanization (Skymask):** Fig. 8 (E) shows the error of G-ICP reached a peak as the skymask changing. Besides, the LOAM error increased in Fig. 8 (F) when the density of the buildings drastically drops.

*b) Verification of the listed three factors using Data2*

**Data2 - Factor 1 - Motion Difference:** As shown in Fig. 9 (A), the translation error jumped to more than 1.0 m for both GICP and Fast LOAM. The ego-vehicle resumed as the traffic light turning back to green from red, and the motion difference increased from 0 to 10 meters. Both methods achieved superior performance around frame 300 between Fig. 9 (A) and (B), as there were a stopping scenario and motion equal to 0. (B) in Fig. 9, the ego car turned to the right through a busy crossroads. The error was up to 1.5 meters due to the motion difference plus the number of dynamic vehicles.

**Data2 - Factor 2 - Dynamic Objects:** The scenes of heavy traffic and numerous cars are shown in Fig. 9 (C)(D). It shows that the dynamic objects have a significant impact on the error.

**Data2 - Factor 3 - Degree of the Urbanization (Skymask):** As shown in Fig. 9 (E)(F) respectively, the rapid changing of skymask resulting in the accuracy loss of the LO methods.

### *D. Discussions*

There are still many open issues for existing LO methods under highly dynamic urban canyons. Table 6 summarizes each LO method's performance, while Table 5 shows the computational efficiency. Given a real-time constrained approach, LOAM provided the most precise and robust performance across the two datasets. Regardless of the computational cost, G-ICP demonstrated the most competitive performance among the point-registration methods on challenging urban scenes.

Another performance comparison of three representative and widely used LO methods, G-ICP, NDT, and LOAM, is shown in Figs. 10 and 11. Considering the impact factors, LOAM is prone to be affected by the large motion difference, as shown in Fig. 10 (A) and Fig. 11 (B) (the motion difference can be seen in Figs. 8 and 9).

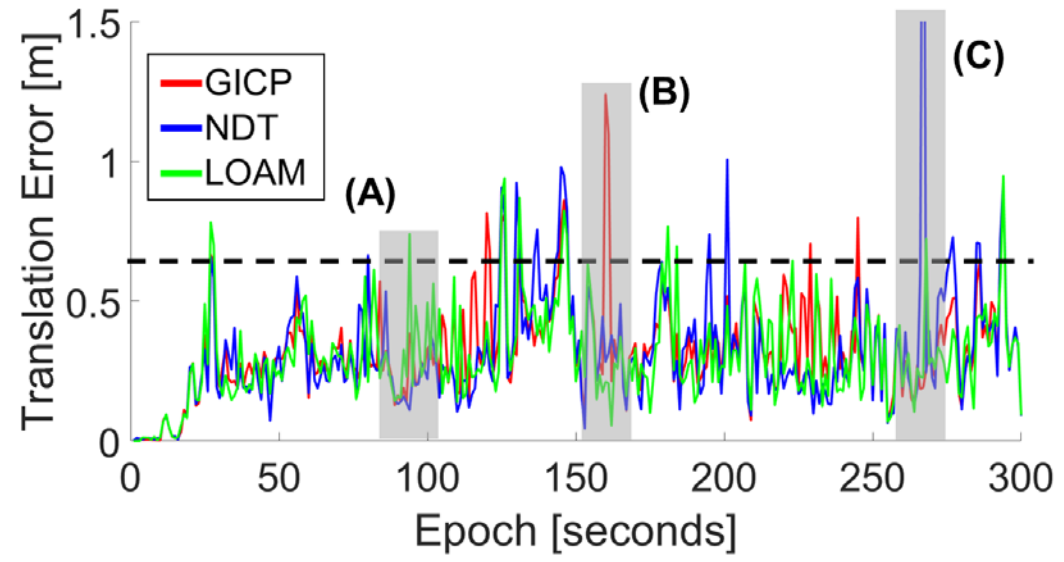

Fig. 10. The performance comparison of G-ICP (red), NDT (blue), and LOAM (green) in Data1, respectively. The dash-dot line indicated the threshold for outliers (~0.7m). (A)(B)(C) and the rectangle are the labels of the performance divergence among the LO methods.

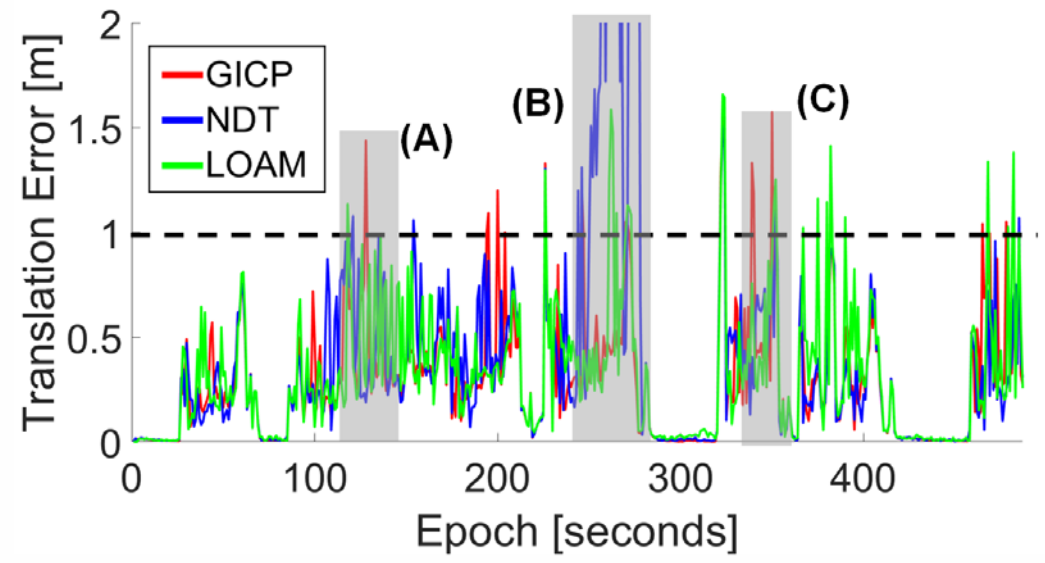

Fig. 11. The performance comparison of G-ICP (red), NDT (blue), and LOAM (green) in Data2, respectively. The dash-dot line indicated the threshold for outliers (~1.0m). (A)(B)(C) and the rectangle are the labels of the performance divergence among the LO methods.

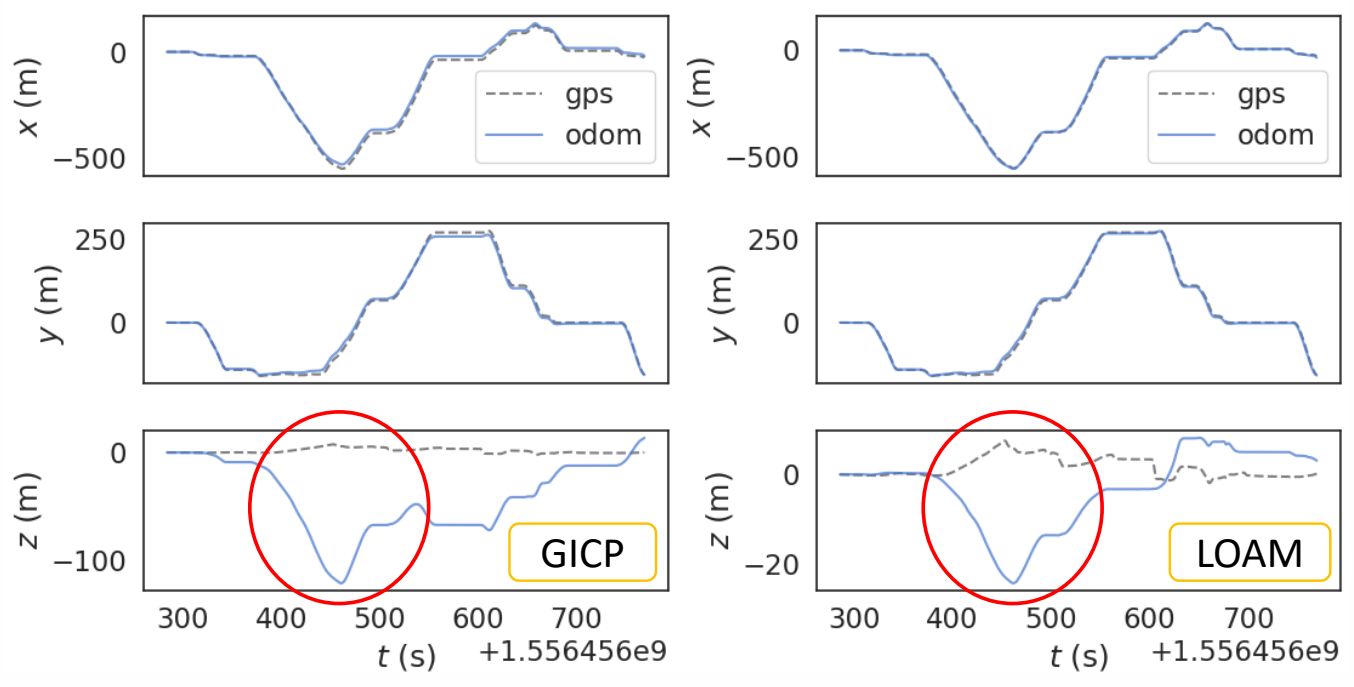

Fig. 12. The trajectory comparison of G-ICP and LOAM versus Ground Truth in Data2 in terms of XYZ-direction. A large difference between LO and ground truth on the z-axis is marked in the red circle.

In numerous dynamic object scenes in Fig. 10 (A)(C), the LOAM and NDT performed not as accurately as G-ICP methods. The performance of G-ICP is more sensitive to skymask change in Fig. 10 (B) and Fig. 11 (A)(C). Therefore, the user can consider a suitable or mix LO approaches to tackle the challenges of different environments in deploying the LO methods for state estimation in a large-scale application in urban canyons.

Furthermore, the evaluation results of Data2 indicate the Z-axis has a larger error comparing to that in XY-directions, as shown in Fig. 12. One possible reason is the horizontal FOV of the Velodyne HDL-32E LiDAR is much wider than that in the vertical direction, e.g., 360°and 40° in horizontal and vertical directions, respectively. It could be challenging to provide state constraints if the vehicle is suffering more up and down movements. As a result, the performance of LiDAR odometry is still challenged in the area with several layers of the viaduct.

## V. CONCLUSIONS AND FUTURE WORK

In this paper, we have presented a benchmark comparison and error analysis of publicly available and popular LO methods based on two challenging datasets collected in urban canyons of Hong Kong. The results are presented in Section. IV and Fig. 13 suggested that the point-wise LO methods can be improved with additional computation load, such as G-ICP. Voxelization from VGICP can reduce the computational load of neighboring point searching but might also lead to accuracy loss in state estimation. Feature-wise methods show both accuracy and cost-efficiency based on the extracted feature points. However, the performance of both methods is affected by the listed three kinds of dominant factors. According to our experiments, we suggest that combining both the feature-wise and point-wise methods can be a promising solution. Firstly, the feature-wise method can efficiently provide a coarse odometry estimation. Then, the coarse one can be applied as an initial guess for the point-wise point cloud registration which relies heavily on the initial guess, leading to a coarse-to-fine LiDAR odometry pipeline. However, it is important to note the feature-wise method might also provide an erroneous estimation. In this case, a fault detector such as the degeneration factor [53] should be used.

In the future, we will study the combination of the point-wise and feature-wise to improve the performance of the existing LOs. Moreover, we will study the degeneration of the LOs in feature insufficient environments and its integration with other sensors, like GNSS and inertial measurement units. We will also investigate the performance of Neural Network (NN) based feature extraction LiDAR odometry as it has the potential for feature extraction and scene segmentation accurately. Last but not least, the performance of LO methods under adverse weather conditions [54, 55] is a good direction to study in urban areas.

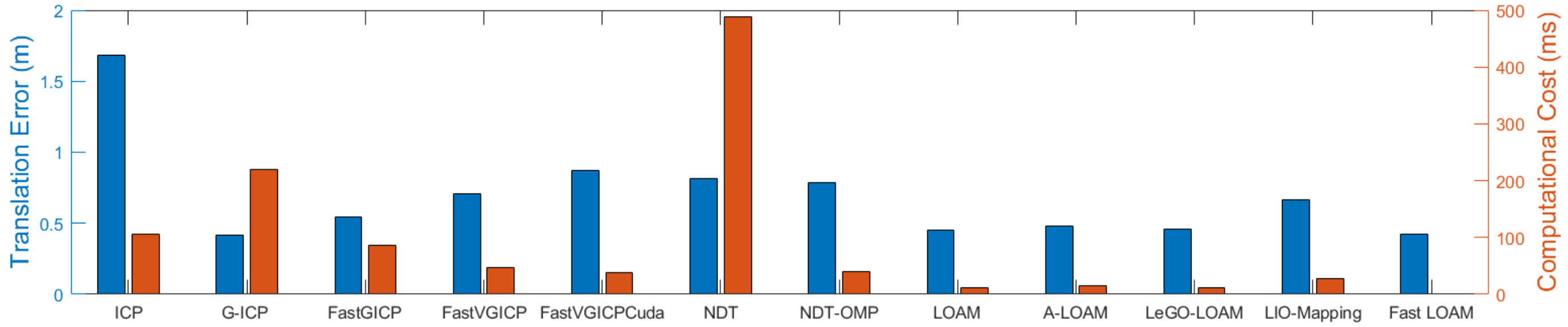

Fig. 13. Comparison of the publicly available LiDAR Odometry algorithms in Data2 in terms of translational error and average computation cost (scan-to-scan).

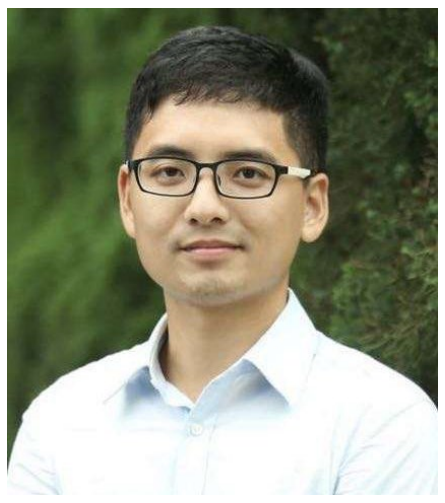

**Feng Huang** (S'21) received his bachelor's degree from Shenzhen University in Automation in 2014 and MSc in Electronic Engineering at Hong Kong University of Science and Technology in 2016. He is a Ph.D. student in the Department of Aeronautical and Aviation Engineering, Hong Kong Polytechnic University. His research interests including localization and sensor fusion for autonomous driving.

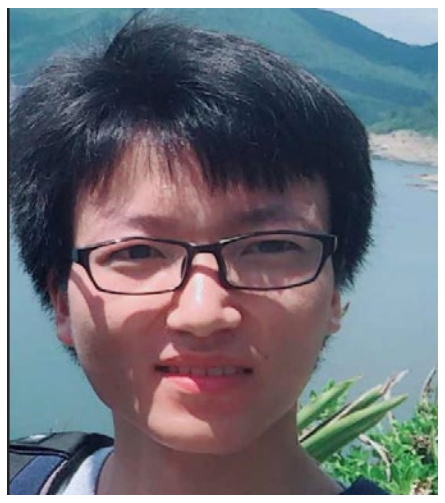

**Weisong Wen** was born in Ganzhou, Jiangxi, China. He is a Ph.D. candidate in mechanical engineering, the Hong Kong Polytechnic University. His research interests include multi-sensor integrated localization for autonomous vehicles, SLAM, and GNSS positioning in urban canyons. He was a visiting student researcher at the University of California, Berkeley (UCB) in 2018.

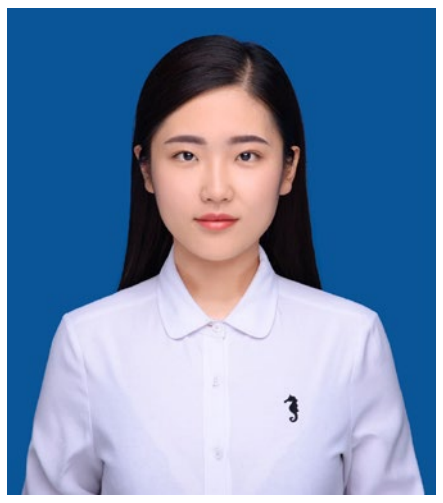

**Jiachen Zhang** received her bachelor's degree from Tianjin University in Information Engineering in 2016 and is currently an enrolled, full-time graduate student at Tianjin University, majoring in Optical Engineering. She is working as a research assistant in the Intelligent Positioning and Navigation Laboratory. Her research interests including localization and sensor-fusion for autonomous driving.

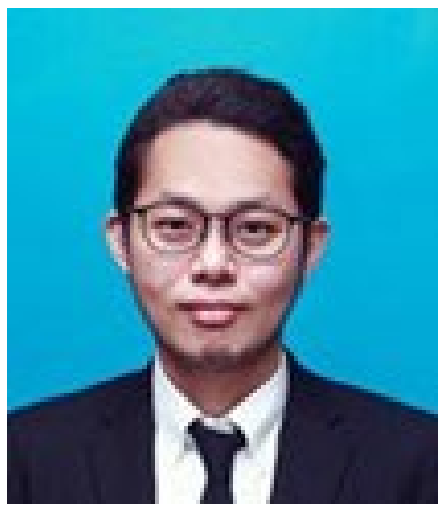

**Li-Ta Hsu** (S'09-M'15) received the B.S. and Ph.D. degrees in aeronautics and astronautics from National Cheng Kung University, Taiwan, in 2007 and 2013, respectively. He is currently an assistant professor with the Department of Aeronautical and Aviation Engineering, The Hong Kong Polytechnic University, before he served as a post-doctoral researcher in the Institute of Industrial Science at the University of Tokyo, Japan. In 2012, he was a visiting scholar in University College London, the U.K. His research interests include GNSS positioning in challenging environments and localization for pedestrian, autonomous driving vehicle, and unmanned aerial vehicle.